\documentclass[10pt,twocolumn,letterpaper]{article}

\usepackage[pagenumbers]{cvpr} 
\usepackage{comment}
\usepackage{amsthm}

\usepackage{xcolor}

\newcommand{\TODO}[1]{\textbf{\color{red}[TODO: #1]}}
\renewcommand{\TODO}[1]{}

\definecolor{cvprblue}{rgb}{0.21,0.49,0.74}
\definecolor{cvprblue}{rgb}{0.21,0.49,0.74}
\usepackage[pagebackref,breaklinks,colorlinks,allcolors=cvprblue]{hyperref}
\usepackage{orcidlink}
\def\confName{CVPR}
\def\confYear{2027}

\title{Not all world models can doubt: how training objective and architecture shape epistemic uncertainty in JEPA and RSSM world models}

\author{
Simon Baur\,\orcidlink{0009-0009-4307-3078} \quad
Wojciech Samek\,\orcidlink{0000-0002-6283-3265} \quad
Jackie Ma\,\orcidlink{0000-0002-2268-1690}\\
Fraunhofer Heinrich-Hertz-Institut, 10587 Berlin, Germany\\
{\tt\small \{simon.baur, wojciech.samek, jackie.ma\}@hhi.fraunhofer.de}
}

\begin{document}
\maketitle
\begin{abstract}
World models predict how an environment responds to actions, so that candidate actions
can be evaluated before execution. Executed actions, or decisions, taken in this manner 
are only as reliable as the model's predictive capabilities, and predictions in turn are only reliable 
within the model's experience. Epistemic uncertainty (EU) is the signal that marks where
that experience ends. 
In this work we offer an extensive empirical study on how training objectives and architectures shape epistemic uncertainty of world models. More precisely, we present a controlled comparison of post-hoc EU (last-layer Laplace and ensembles) between two dominant models of different paradigms: 1) LeWorldModel (LeWM): A joint-embedding predictive architecture (JEPA), which predicts the next embedding without a decoder, and 2) DreamerV3: A recurrent state-space model (RSSM), which reconstructs the observation from a recurrent latent state.  In ten simulated visual-control environments, we follow a controlled state perturbation of the environment state through every stage of the models up to the final EU estimate. We find that the RSSM attenuates the state perturbation through two mechanisms before it reaches an estimator on the predictor feature. Its recurrent update averages over context, so a change in a single frame reaches the state at a fraction of its size and recovers only as the change persists. Its reconstruction objective weights an entity by the pixels it covers, so a small object can be lost due to the training objective. The JEPA has neither mechanism and passes the change through all stages to the uncertainty. We further show that this translates to gaps in five uncertainty related downstream tasks, from unfamiliar-state detection to uncertainty-penalised planning: the RSSM uncertainty misses deviations that its own posterior registers, whereas the JEPA uncertainty reports what its model sees.

\end{abstract}

\section{Introduction}
\label{sec:intro}

\begin{figure*}[t]
  \centering
   \includegraphics[width=1.0\linewidth]{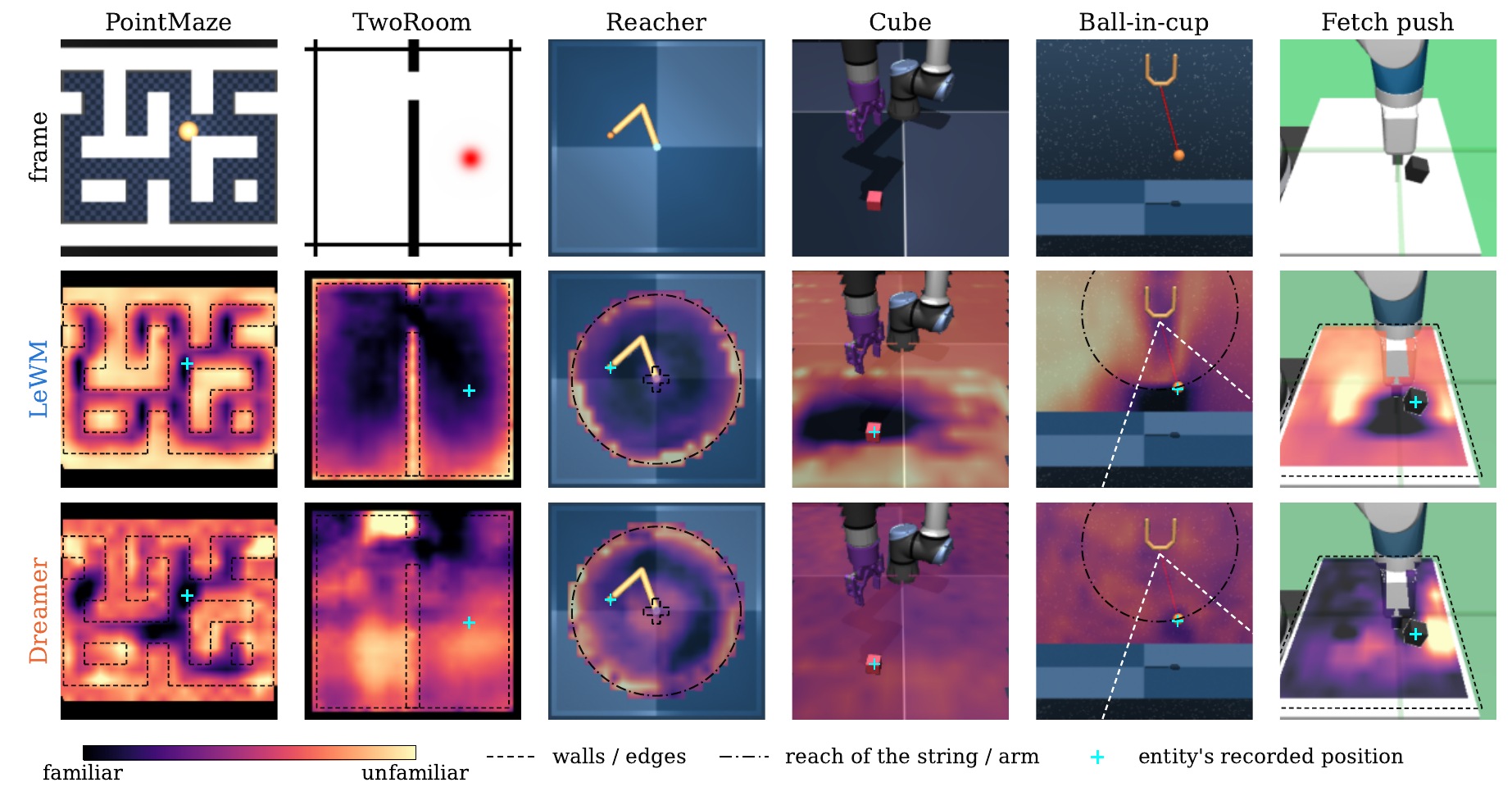}
   \caption{\textbf{Uncertainty landscapes.} The controlled entity is placed at every point of a grid over its surface, and each model's $u$ uncertainty is displayed at that point (computation details in Appendix D). For LeWM  uncertainty is low where the entity can be, rises at walls (dashed) and reach limits (dash-dotted), and grows gradually with displacement from the recorded position. Dreamer is nearly flat or varies erratically, with no consistent relation to either environment dynamic.}
   \label{fig:landscapes}
\end{figure*}

World models, learned simulators that predict how an environment responds to an actors actions, have become a central component of embodied artificial intelligence~\citep{ha2018recurrent, lecun2022path, hafner2025mastering}. They let an actor evaluate candidate actions in imagination, from image or video data, without executing them: examples include zero-shot robot planning with large scale model pretrained models~\citep{assran2025v}, manipulation and navigation planned in the latent space of a model trained on small task-specific datasets~\citep{zhou2024dino, sobal2026learning, maes2026leworldmodel}, driving in dense traffic~\citep{henaff2019model}, and long-horizon control through hierarchical latent planning~\citep{zhang2026hierarchical}. In all these settings the agent acts on imagined rather than observed outcomes, so its reliability is bounded by that of the model, and a learned model is reliable only within the support of its training data; beyond it, a planner optimising against the model is subject to its errors~\citep{janner2019trust, lambert2020objective, bhamidipaty2026imperfect}. Epistemic uncertainty, an estimate of the model's lack of experience at a given state, is the signal that separates predictions to act on from predictions to avoid, and it underlies exploration, offline policy optimisation and safety filtering in model-based control~\citep{sekar2020planning, yu2020mopo, seo2025uncertainty}. For world models learned from image or video data it has received little attention, and the available studies take a critical stance: estimates derived from the latent space can be systematically biased~\citep{becker2022uncertainty, berger2026biased}, and model uncertainty might not reliably function as a risk signal~\citep{wang2026learning}. World models differ most consequentially in what they are trained to predict, and different training paradigms emerged. Recurrent state-space models (RSSMs), e.g. Dreamer~\citep{hafner2025mastering}, optimise a variational bound in which a stochastic latent reconstructs the observation and a recurrent state summarises the history. Reconstruction provides a dense training signal, a generative model that renders its own imagined futures and a stochastic latent for aleatoric variability, and the resulting agents master a broad range of control tasks from pixels~\citep{hafner2025mastering}; its cost is that capacity follows pixel variance rather than task relevance, so that visually dominant but behaviourally irrelevant content is modelled first~\citep{balestriero2024learning}. Joint-embedding predictive architectures (JEPA) omit the decoder and predict the next embedding from the current embeddings and actions under a regulariser against collapse~\citep{lecun2022path, assran2023self, maes2026leworldmodel}. Latent prediction allocates capacity to what is predictable~\citep{van2026joint}, scales to internet video~\citep{assran2025v} and supports stable end-to-end training and planning from pixels~\citep{maes2026leworldmodel}, at the cost of a deterministic predictor without a native account of stochasticity and a latent space that cannot be inspected by decoding; probabilistic formulations are only beginning to appear~\citep{huang2026vjepa, gogl2026var, song2026branchjepafinitesupportpredictivedistributions}. For neither paradigm is epistemic uncertainty well understood yet. In RSSMs it has been estimated with ensembles of transition heads, whose disagreement misses the error that accumulates over latent rollouts~\citep{sekar2020planning, berger2026biased}; for JEPA world models post-hoc uncertainty has not been studied, the closest work deriving a density score from a frozen JEPA encoder for outlier detection in images~\citep{balestriero2025gaussian}. Last-layer Bayesian approximations, of which the Laplace approximation is the most common, are post hoc, inexpensive and well understood~\citep{mackay1992practical, kristiadi2020being, daxberger2021laplace}, but their estimates are functions of the penultimate features: a variation of the input that does not move the features is invisible to any estimator built on them~\citep{van2021feature, wilson2026last}. In a world model these features are produced by an encoder, a dynamics stage and, in recurrent models, a state that integrates the context, none of which is trained to preserve the variations an uncertainty estimate must detect. How much of a change in the environment state survives to the point where the uncertainty is read, and how this depends on objective and architecture, therefore decides whether such estimates are useful at all. Our work aims to provide this analysis. We trace a controlled state perturbation through the stages of each paradigm, locate where and how representations influence uncertainty estimates, and evaluate how these translate to related downstream tasks. 
\begin{figure*}[!h]
  \centering
   \includegraphics[width=1.0\linewidth]{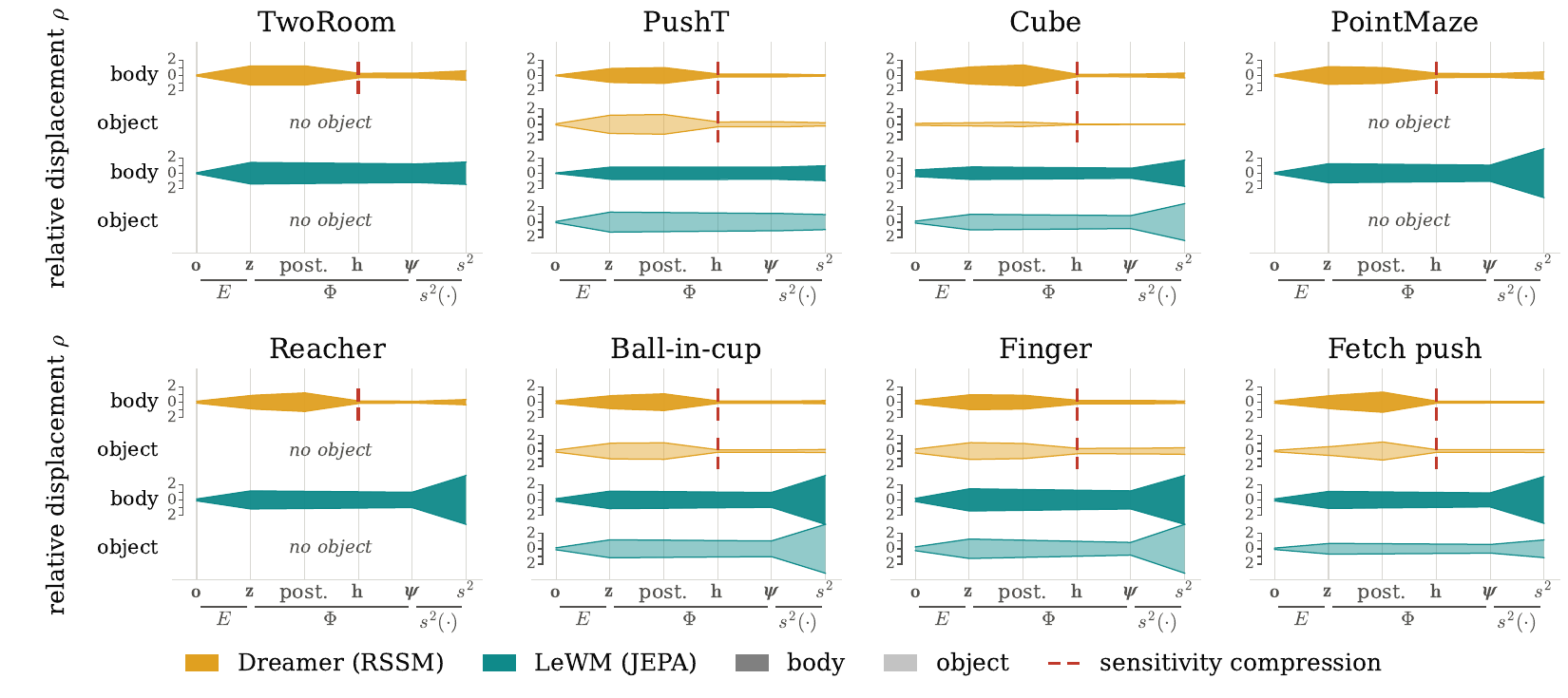}

\caption{\textbf{Stage trace of a state perturbation.} Relative displacement $\rho$ at each stage of Eq.~\eqref{eq:chain} for a perturbation of the body (dark) and the object (light), ending in the sensitivity $\kappa$ of the uncertainty. Both encoders transmit the perturbation. Dreamer attenuates it at the GRU update (dashed marker) and the later stages inherit attenuation; LeWM passes it through to $\boldsymbol{\psi}$ and $s^2$. Scene and Cube double are omitted due to space, Appendix C contains quantities of this analysis for all 10 environments.}
   \label{fig:trace_analysis}
\end{figure*}

Specifically, we \begin{itemize}[leftmargin=*, itemsep=2pt]
  \item present 
  the first controlled comparison of post-hoc epistemic uncertainty between a JEPA (LeWM~\citep{maes2026leworldmodel}) and a reconstruction-based RSSM (DreamerV3~\citep{hafner2025mastering}), with identical estimator, data and evaluation protocol across 10 simulated visual-control environments;
  \item trace controlled state perturbations from the observations to the final uncertainty estimate and diagnose two attenuation mechanisms for the RSSM case: the recurrent update, which admits a one-frame change only in part, and the reconstruction objective, which weights entities by pixel area, diluting changes to small objects; JEPA however preserves the response at every stage;
  \item evaluate both paradigms on five downstream tasks: unfamiliar-state detection, error ranking, physical-violation detection, response to consistent distribution shifts and uncertainty-penalised planning, and find that the results follow the diagnosis: the JEPA uncertainty reports what its model registers, including single-frame violations of physics and the physical structure of the environment, whereas the RSSM uncertainty misses deviations that are brief or concern small entities.

\end{itemize}

\section{Related Work}
\label{sec:related}

\paragraph{Post-hoc uncertainty on frozen features.}
Last-layer Bayesian approximations treat the network up to its penultimate layer as a fixed feature map and place a posterior on the final linear layer alone. The Laplace approximation~\citep{mackay1992practical} is the most common instance, fitted after training in a single pass over the data, and removes much of the overconfidence of deterministic networks~\citep{daxberger2021laplace, kristiadi2020being, ritter2018scalable}; deep ensembles~\citep{lakshminarayanan2017simple} are the main alternative, with a post-hoc variant that trains several heads on shared frozen features~\citep{osband2016deep, wen2020batchensemble}. Both report the distance of a test feature from the training features, so their quality is bounded by the feature map~\citep{van2021feature, mukhoti2023deep, wilson2026last}. In world models, ensembles are applied in this post-hoc fashion, as several heads on the features of a trained model~\citep{sekar2020planning, berger2026biased}.

\paragraph{Epistemic uncertainty in world models.}
In model-based reinforcement learning the epistemic uncertainty of the dynamics model is almost always an ensemble of transition models~\citep{chua2018deep, kurutach2018model, janner2019trust}, used to drive exploration~\citep{pathak2019self, sekar2020planning}, to penalise the reward in offline optimisation~\citep{yu2020mopo, kidambi2020morel, rafailov2021offline} or to filter unsafe actions~\citep{kahn2017uncertainty, henaff2019model, seo2025uncertainty}, on the assumption that disagreement tracks the model's lack of experience. For latent-state models trained from pixels this assumption has been questioned: disagreement in a learned latent space can be systematically biased~\citep{becker2022uncertainty}, misses the error that accumulates along imagined rollouts~\citep{berger2026biased} and does not reliably predict failure of the resulting policy~\citep{wang2026learning}. These studies, however, assess the uncertainty at its output and leave open how it is formed in the first place. In this study, we trace the uncertainty estimate backwards through the stages of the model and ask what the influencing factors are. Furthermore, we extend the question from the RSSM setting to the JEPA paradigm.

\paragraph{Reconstruction and latent-prediction objectives.}
An RSSM~\citep{hafner2019learning, hafner2019dream, hafner2025mastering} minimises a variational bound on the likelihood of the observations, $\mathcal L_{\mathrm{RSSM}} = \sum_t \mathbb E_{q}[-\log p(\mathbf o_t \mid \mathbf h_t,\mathbf s_t)] + \beta\,\mathrm{KL}[q_t \,\|\, p_t]$, with $q_t = q(\mathbf s_t\mid\mathbf h_t,\mathbf z_t)$ the posterior and $p_t = p(\mathbf s_t\mid\mathbf h_t)$ the prior over the latent state (Sec.~\ref{sec:chain}).The first term is a per-pixel reconstruction loss, known to spend capacity in proportion to pixel variance rather than task relevance~\citep{balestriero2024learning}; a line of work removes it from the RSSM~\citep{okada2021dreaming, nguyen2021temporal, deng2022dreamerpro, wang2022denoised} or adds object-centric structure to protect small entities~\citep{ferraro2025focus}. A JEPA~\citep{lecun2022path, assran2023self, bardes2024revisiting} replaces reconstruction by prediction in embedding space, $\mathcal L_{\mathrm{JEPA}} = \sum_t \|\Phi(\mathbf z_{\le t}, \mathbf a_{\le t}) - \operatorname{sg}[\mathbf z_{t+1}]\|^2 + \mathcal R(\mathbf z)$, with $\mathbf z_{\le t}$ and $\mathbf a_{\le t}$ the last $H$ embeddings and actions (Sec.~\ref{sec:chain}), $\operatorname{sg}$ a stop-gradient on the target and $\mathcal R$ a regulariser against collapse~\citep{bardes2021vicreg}.

\paragraph{Physical understanding in reconstruction and embedding models.}
Whether a world model represents the physics of a scene is usually tested by violation of expectation: the model is shown a physically impossible event and asked whether it is surprised. Pixel-space generative video models pass such tests only partially~\citep{kang2024far, motamed2026generative}, whereas JEPA models trained on natural video register violations of object permanence, continuity and solidity from their prediction error in embedding space~\citep{garrido2025intuitive}, consistent with earlier evidence that object-centric representations predict physical outcomes better than pixel-based ones~\citep{bear2021physion}. These studies read the prediction error, which requires the true next frame. We ask the complementary question for the uncertainty, which does not: whether a post-hoc epistemic estimate inherits the physical sensitivity of the representation it sits on.

\section{Analytical framework}
\label{sec:analyitcal_framework}
\begin{figure}[!h]
  \centering
   \includegraphics[width=1.0\linewidth]{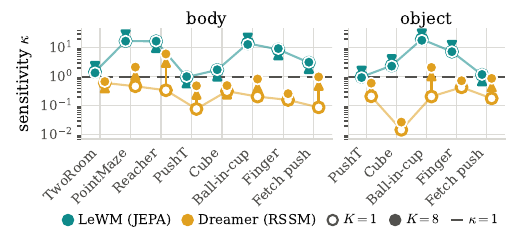}
\caption{\textbf{Sensitivity per environment.} $\kappa$ for body and object perturbations. LeWM is sensitive throughout; Dreamer attenuates every perturbation at $K=1$ and recovers partly with $K=8$.}
   \label{fig:kappa}
\end{figure}

\subsection{Post-hoc uncertainty from predictor features}
\label{sec:record}

\label{sec:record}

\begin{figure*}[!h]
  \centering
   \includegraphics[width=1.0\linewidth]{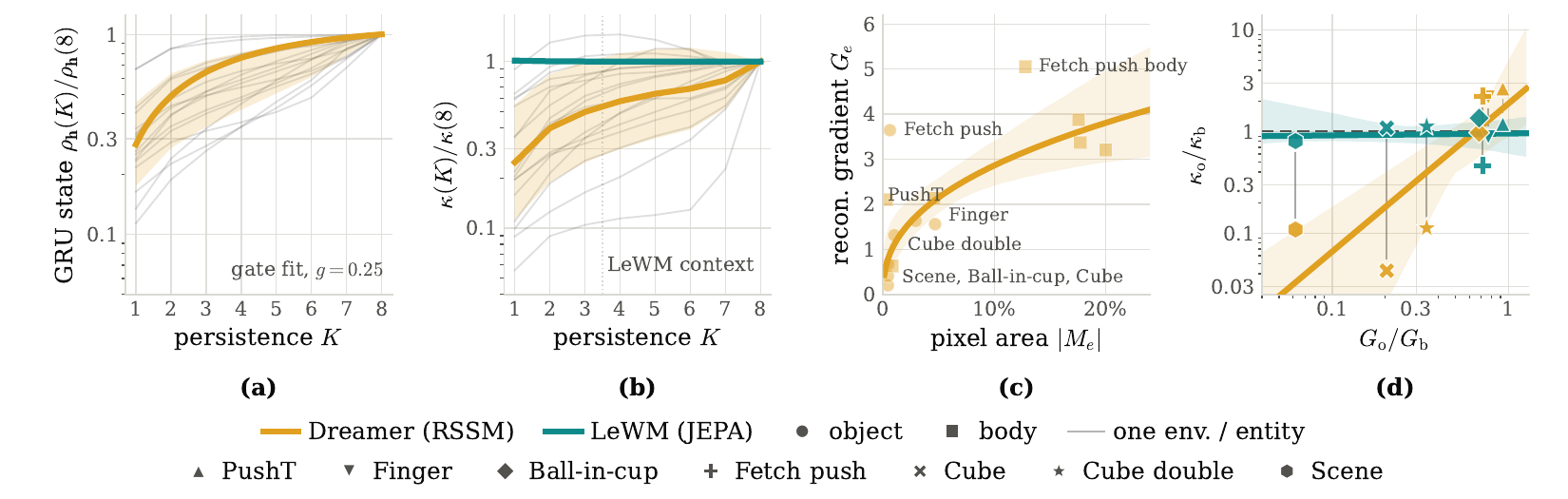}
\caption{\textbf{Mechanistic influences of architectural and paradigm factors on uncertainty sensitivity.} Architecture: the longer a perturbation persists, the more of it Dreamer's recurrent state retains \textbf{(a)} and the more sensitive its uncertainty becomes \textbf{(b)}; LeWM is independent of persistence. Paradigm: Dreamer's reconstruction gradient grows with the pixels an entity covers \textbf{(c)}, and the entity's share of it sets the sensitivity to that entity \textbf{(d)}; LeWM is equally sensitive to body and object.}
\label{fig:mech-factors}
\end{figure*}
\subsection{Stage-wise sensitivity to state perturbations}
\label{sec:trace}

Let a world model be trained and frozen, and let its predictor end in a linear layer with weight matrix $W \in \mathbb{R}^{D \times F}$. For an observation window $\mathbf{w}$ we write $\boldsymbol{\psi}(\mathbf{w}) \in \mathbb{R}^{F}$ for the input of that layer, the \emph{predictor feature}, and $\hat{\mathbf{y}} = W\boldsymbol{\psi}$ for the predicted next latent. A \emph{post-hoc} uncertainty estimator is a map
\begin{equation}
  u(\boldsymbol{\psi}) \;=\; U\big(\boldsymbol{\psi};\, \mathcal{D}\big),
  \qquad
  \mathcal{D} = \{\boldsymbol{\psi}_n\}_{n=1}^{N},
  \label{eq:posthoc}
\end{equation}
fitted on the predictor features of the training windows and depending on a test window only through $\boldsymbol{\psi}$.

\paragraph{Last-layer Laplace.}
Under a Gaussian likelihood with noise $\sigma^2$ and an isotropic Gaussian prior with variance $\tau^2$ on the rows of $W$, the posterior over $W$ is Gaussian with a precision shared across rows,
\begin{equation}
  \Lambda \;=\; \frac{1}{\sigma^2}\, G + \frac{1}{\tau^2}\, I,
  \qquad
  G \;=\; \sum_{n=1}^{N} \boldsymbol{\psi}_n \boldsymbol{\psi}_n^{\top},
  \label{eq:precision}
\end{equation}
and the epistemic predictive variance at a feature vector is the quadratic form
\begin{equation}
  s^2(\boldsymbol{\psi}) \;=\; \boldsymbol{\psi}^{\top} \Lambda^{-1} \boldsymbol{\psi}
  \;=\; \sum_{i=1}^{F} \frac{\langle \boldsymbol{\psi}, \mathbf{e}_i\rangle^2}{\lambda_i/\sigma^2 + 1/\tau^2},
  \label{eq:laplace}
\end{equation}
with $(\lambda_i, \mathbf{e}_i)$ the eigenpairs of the Gram matrix $G$, accumulated in one pass over the training windows. The subsequent analysis is conducted with $u = s^2$ as in Eqs.~\eqref{eq:precision} and~\eqref{eq:laplace}. Appendix C provides an estimator ablation with an ensemble of predictor heads, and observes no material difference ($r \ge 0.98$ of across task scores). Since $u$ is a function of $\boldsymbol{\psi}$ alone, it can register a perturbation of the environment state only if $\boldsymbol{\psi}$ does. Which perturbations $\boldsymbol{\psi}$ preserves in turn is determined by the model architecture and the training objective, and the subject of the next section. We contrast the two prevailing \emph{paradigms}, joint-embedding prediction and reconstruction-based recurrent state-space models, with LeWM as the JEPA instance and DreamerV3 as the RSSM instance; where we write RSSM in the following, we refer to the architecture of PlaNet and Dreamer, that is, a gated recurrent update trained with a reconstruction term.

\subsection{Composition of $\boldsymbol{\psi}$ in JEPA and RSSMs}
\label{sec:chain}
The predictor feature is a mapping $\boldsymbol{\psi}_t = F_\theta(\mathbf{x}_{t-H+1:t})$ of the environment states $\mathbf{x}\in \mathbb{R}^{n}$ of the window $\mathbf{w}$, with parameters $\theta$ fixed by training. With $R$ the renderer producing the observation $\mathbf{o}$, $E$ the encoder and $\Phi$ the map from the embeddings of a window to the predictor feature, $F_\theta$ composes in both architectures as the chain
\begin{equation}
  \mathbf{x} \xrightarrow{R} \mathbf{o} \xrightarrow{E} \mathbf{z}
  \xrightarrow{\Phi} \boldsymbol{\psi} \xrightarrow{U} u ,
  \label{eq:chain}
\end{equation}
whose steps we call \emph{stages} (formal definition in Appendix~\ref{app:chain}). The stages $\mathbf{o}$, $\mathbf{z}$, $\boldsymbol{\psi}$ and $u$ exist in both paradigms; they differ in $\Phi$. In the JEPA case, $\Phi$ is a feed-forward predictor over the last $H$ embeddings $\mathbf{z}_t = E(\mathbf{o}_t)$ and actions, $\boldsymbol{\psi}_t = \Phi(\mathbf{z}_{t-H+1:t}, \mathbf{a}_{t-H+1:t})$ with $\hat{\mathbf{z}}_{t+1} = W\boldsymbol{\psi}_t$, so every frame of the window enters $\boldsymbol{\psi}_t$ directly and nothing is carried over from earlier frames. In the RSSM case, $\Phi$ contains two further stages: a posterior $\mathbf{q}_t = q(\mathbf{z}_t, \mathbf{h}_t)$ over a categorical state is inferred from the embedding and the recurrent state, a sample $\mathbf{s}_t \sim q(\mathbf{s}_t \mid \mathbf{q}_t)$ advances the recurrent state through a GRU, $\mathbf{h}_{t+1} = \mathrm{GRU}([\mathbf{s}_t, \mathbf{a}_t], \mathbf{h}_t)$, and the prior network maps the new state to the feature $\boldsymbol{\psi}_{t+1} = \Phi_{\mathrm{prior}}(\mathbf{h}_{t+1})$ from which the next state is predicted, $\mathbf{s}_{t+1} \sim p(\mathbf{s}_{t+1} \mid W\boldsymbol{\psi}_{t+1})$, with $q$ and $p$ categorical. A frame enters $\boldsymbol{\psi}_{t+1}$ only through the GRU update, which also carries information from frames before the window.

\paragraph{State, body and object.}
The environment state $\mathbf{x}$ is the vector of physical coordinates the simulator integrates, e.g.\ positions, angles and velocities. At every step the simulator receives an action $\mathbf{a} \in \mathbb{R}^{m}$, the motor command of the agent, and advances the state by its transition function $\mathcal{T}$. The coordinates fall into two groups,
\begin{align}
  \mathbf{x} &= (\mathbf{x}_{\mathrm{b}}, \mathbf{x}_{\mathrm{o}}),
  \label{eq:body_object} \\
  (\mathbf{x}_{\mathrm{b},t+1}, \mathbf{x}_{\mathrm{o},t+1})
    &= \big(\mathcal{T}_{\mathrm{b}}(\mathbf{x}_t, \mathbf{a}_t),\, \mathcal{T}_{\mathrm{o}}(\mathbf{x}_t)\big).
  \label{eq:transition}
\end{align}
The \emph{body} $\mathbf{x}_{\mathrm{b}}$ is the part of the state the action acts on, the agent itself; the \emph{object} $\mathbf{x}_{\mathrm{o}}$ is the part the action does not act on, which changes only through the current body state, by contact or causality within the environment. Which coordinates exist and which of them the action enters is therefore a property of the dynamics of the environment.

\paragraph{Relative displacement and uncertainty sensitivity under state perturbation.}
Let $\sigma_i$ be the standard deviation of coordinate $i$ of $\mathbf{x}$ over the training data and $\mathbf{e}_i$ its unit vector. We write
\begin{equation}
  \mathbf{x}' \;=\; \mathbf{x} + \epsilon\,\sigma_i\,\mathbf{e}_i
  \label{eq:displaced}
\end{equation}
for the state in which coordinate $i$ is displaced by a fraction $\epsilon$ of its typical variation. For base states $\mathbf{x}$ from unseen episodes, we set the simulator to $\mathbf{x}'$, re-render the current frame and pass the window through the frozen model; the perturbation in that sense is a (impossible) discontinuity of environment physics, the same displacement that Sec.~\ref{sec:tasks} applies as the teleport violation. For any stage output $\mathbf{y}$ of Eq.~\eqref{eq:chain}, the \emph{relative displacement}
\begin{equation}
  \rho_{\mathbf{y}}
  \;=\; \mathbb{E}_{\mathbf{x}}\!\left[
  \frac{\|\mathbf{y}(\mathbf{x}') - \mathbf{y}(\mathbf{x})\|}
       {\epsilon\,\|\mathbf{y}(\mathbf{x})\|}\right]
  \label{eq:rho}
\end{equation}
measures the fraction of one typical variation of the coordinate that is present at that output. The profile $\rho_{\mathbf{o}}, \rho_{\mathbf{z}}, \rho_{\boldsymbol{\psi}}, \rho_{u}$, with the RSSM internals $\rho_{\mathbf{q}}$ and $\rho_{\mathbf{h}}$, is the \emph{stage trace} of a group: a stage whose output displacement falls below its input displacement attenuates the perturbation, and a perturbation removed at any stage is absent from all later ones. The last entry we call \emph{sensitivity} of the uncertainty,
\begin{equation}
  \kappa \;=\; \rho_{u}
  \;=\; \mathbb{E}_{\mathbf{x}}\!\left[
  \frac{\big|u\big(\boldsymbol{\psi}(\mathbf{x}')\big)
  - u\big(\boldsymbol{\psi}(\mathbf{x})\big)\big|}
  {\epsilon\,u\big(\boldsymbol{\psi}(\mathbf{x})\big)}\right],
  \label{eq:kappa}
\end{equation}
which is one when a typical variation of the coordinate changes the uncertainty by its own magnitude and zero when the uncertainty does not respond.

\section{Measured sensitivity of post-hoc uncertainty in JEPA and RSSM}
\begin{figure*}[!h]
  \centering
   \includegraphics[width=1.0\linewidth]{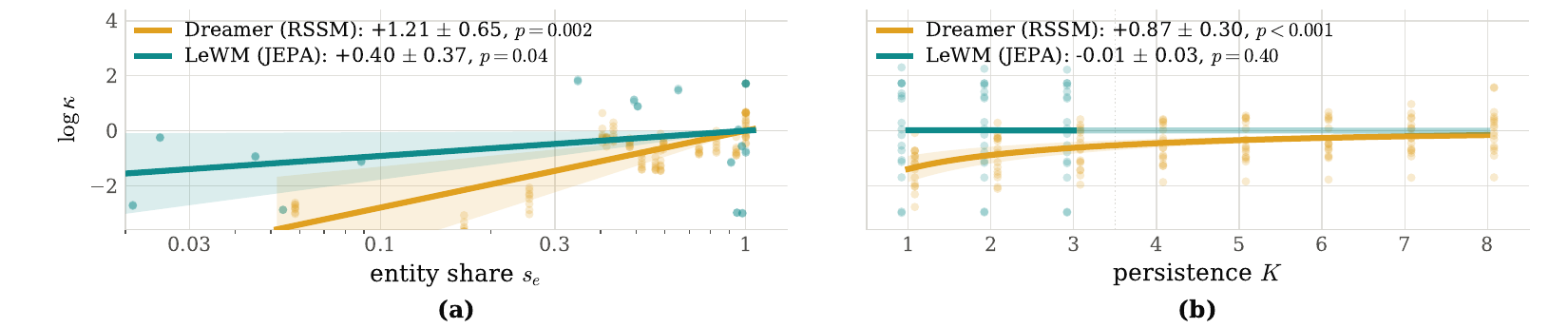}
\caption{\textbf{Persistence and entity share as predictors of sensitivity.} Sensitivity against the entity's share of the training signal \textbf{(a)} and the persistence of the perturbation \textbf{(b)}, each with the other factor accounted for. For Dreamer both factors are significant and together explain two thirds of the variation; for LeWM persistence has no effect and entity share a weak one. LeWM is fitted on its own context, $K \le 3$.}
\label{fig:regression}
\end{figure*}
\label{sec:sensitivity_results}
LeWM and DreamerV3 are trained on the same recorded data per environment and frozen; the Laplace estimate is fitted post hoc on the predictor's last layer. We evaluate on ten environments: TwoRoom, PointMaze, Reacher, PushT, Cube, Ball-in-cup, Finger, Fetch push, Scene and Cube double \citep{chi2025diffusion,tassa2018deepmind,park2025ogbench, plappert2018multi, sobal2026learning}. Figure~\ref{fig:landscapes} shows example uncertainty maps for both models. A state perturbation can enter the context in a single frame or be held over its last $K$ frames; we call $K$ the \emph{persistence} of the perturbation.
Figure~\ref{fig:trace_analysis} shows the stage trace at $K=1$ and Figure~\ref{fig:kappa} the resulting sensitivity per environment and group; Appendix~\ref{sec:appendix_addtional_results} tabulates the effect of increasing persistence $K$. Both encoders transmit the perturbation, $\rho_{\mathbf{z}} \approx 1$ for LeWM and for the Dreamer encoder and posterior, for body and object alike. The chains diverge thereafter. In LeWM the perturbation reaches the predictor feature and is amplified at the uncertainty, $\rho_{\boldsymbol{\psi}} \approx 0.9$ and $\kappa \ge 0.94$ in every environment, with $\kappa$ between $7$ and $20$ in PointMaze, Reacher, Ball-in-cup and Finger. In Dreamer it is attenuated at a single stage, the GRU update, $\rho_{\mathbf{h}} \approx 0.2$ after $\rho_{\mathbf{q}} \approx 1$, and the later stages inherit this value, $\rho_{\boldsymbol{\psi}} \approx 0.2$ and $\kappa < 1$ in all environments and both groups. The Laplace estimate is thus subject to the previous attenuation. Holding the perturbation over $K=8$ frames can restore Dreamer's sensitivity (Figure~\ref{fig:kappa}).
We identify two sources of the attenuation, one in the architecture and one in the training paradigm (Figure~\ref{fig:mech-factors}).

\paragraph{Architecture: the recurrent update averages over the context.}
The RSSM carries its state through a gated recurrent unit, $\mathbf{h}_{t+1} = (1-\mathbf{g}) \odot \mathbf{h}_t + \mathbf{g} \odot \tilde{\mathbf{h}}_t$, whose update gate $\mathbf{g}$ sets how much of the current input enters the state. A perturbation present in a single frame reaches $\mathbf{h}$ through one update, scaled by $\mathbf{g}$, whereas a perturbation held over $K$ frames accumulates. The traces bear this out: $\rho_{\mathbf{h}}$ rises monotonically with $K$ in every environment and group (Figure~\ref{fig:mech-factors}a), by a factor between $2.3$ and $4.6$ from $K=1$ to $K=8$ in eleven of them, and the sensitivity follows (Figure~\ref{fig:mech-factors}b). The GRU thus acts as a moving average over its context, admitting a one-frame perturbation at a fifth of its size and a persistent one at full size. LeWM has no such state; its predictor reads the context embeddings directly, and its sensitivity is independent of $K$.\paragraph{Paradigm: the reconstruction term weights an entity by the pixels it covers.}
The second source is determined by the training objective and concerns the object. The reconstruction term of the ELBO sums the squared error over pixels, so an entity enters the objective in proportion to the pixels it covers. That this weighting lets reconstruction-based world models neglect small but task-relevant objects is known~\citep{okada2021dreaming, deng2022dreamerpro, seo2023masked}; we show that it also governs how a perturbation of the object propagates to the uncertainty. Restricting the reconstruction term to the pixels of an entity $e \in \{\mathrm{b}, \mathrm{o}\}$ and taking the norm of its gradient with respect to the state the decoder reads, $G_e$, measures the strength with which reconstruction pulls the state towards representing that entity, the quantity the ELBO trades against the KL cost of keeping its coordinates (Appendix B). The gradient grows with the pixel area of the entity (Figure~\ref{fig:mech-factors}c; Spearman $0.65$), and in Dreamer the entity's share of it carries through to the uncertainty: $\kappa^{\mathrm{o}}/\kappa^{\mathrm{b}}$ at $K=1$ follows $G_{\mathrm{o}}/G_{\mathrm{b}}$ across the seven object environments (Figure~\ref{fig:mech-factors}d; Spearman $0.93$), from below $0.1$ where the object draws a tenth of the body's gradient to above $1$ where it draws as much. The JEPA objective has no such term, and LeWM keeps $\kappa^{\mathrm{o}} \approx \kappa^{\mathrm{b}}$ in the same environments (Spearman $0.21$). 

\paragraph{Joint account.}
A regression of $\log\kappa$ on the entity's gradient share and a persistence term over all environment--entity--$K$ cells (Figure~\ref{fig:regression} and Appendix B) confirms that the two factors act jointly. For Dreamer both are significant, gradient share $+1.21 \pm 0.65$ and persistence $+0.87 \pm 0.30$, and together explain two thirds of the variation ($R^2 = 0.65$). For LeWM, fitted on its own context $K \le 3$, persistence has no effect ($-0.01 \pm 0.03$) and entity share a weak one ($+0.40 \pm 0.37$), a third of Dreamer's. The sensitivity of the JEPA model is thus set by neither factor; that of the RSSM is set by both.

\section{Downstream tasks}
\begin{table}[!h]
\centering
\small
\caption{\textbf{Unfamiliar-state detection.} AUROC of $s^2$ for states from a region withheld when fitting the Laplace head, on the body and the object factor. Mean and sd over four withheld quadrants.}
\label{tab:detection_main}
\begin{tabular}{lcccc}
\toprule
& \multicolumn{2}{c}{body} & \multicolumn{2}{c}{object} \\ \cmidrule(lr){2-3} \cmidrule(lr){4-5}
Environment & LeWM & Dreamer & LeWM & Dreamer \\
\midrule
TwoRoom & \textbf{0.96}{\scriptsize$\pm$.02} & 0.89{\scriptsize$\pm$.02} & -- & -- \\
PointMaze & \textbf{1.00}{\scriptsize$\pm$.00} & 0.98{\scriptsize$\pm$.02} & -- & -- \\
Reacher & \textbf{0.84}{\scriptsize$\pm$.02} & 0.80{\scriptsize$\pm$.02} & -- & -- \\
PushT & \textbf{0.96}{\scriptsize$\pm$.01} & 0.73{\scriptsize$\pm$.02} & \textbf{0.98}{\scriptsize$\pm$.01} & 0.80{\scriptsize$\pm$.05} \\
Cube & \textbf{0.97}{\scriptsize$\pm$.00} & 0.87{\scriptsize$\pm$.03} & \textbf{0.99}{\scriptsize$\pm$.00} & 0.87{\scriptsize$\pm$.02} \\
Ball-in-cup & \textbf{0.98}{\scriptsize$\pm$.01} & 0.94{\scriptsize$\pm$.03} & \textbf{0.98}{\scriptsize$\pm$.00} & 0.93{\scriptsize$\pm$.03} \\
Finger & 0.81{\scriptsize$\pm$.10} & \textbf{0.88}{\scriptsize$\pm$.07} & 0.78{\scriptsize$\pm$.03} & \textbf{0.84}{\scriptsize$\pm$.01} \\
Fetch push & \textbf{0.97}{\scriptsize$\pm$.00} & 0.90{\scriptsize$\pm$.03} & \textbf{0.97}{\scriptsize$\pm$.00} & 0.90{\scriptsize$\pm$.03} \\
Scene & 0.67{\scriptsize$\pm$.02} & \textbf{0.75}{\scriptsize$\pm$.05} & \textbf{0.54}{\scriptsize$\pm$.03} & 0.48{\scriptsize$\pm$.08} \\
Cube double & 0.76{\scriptsize$\pm$.05} & \textbf{0.95}{\scriptsize$\pm$.03} & \textbf{0.97}{\scriptsize$\pm$.00} & 0.67{\scriptsize$\pm$.04} \\
\bottomrule
\end{tabular}
\end{table}
\label{sec:tasks}

\begin{figure*}[!h]
  \centering
   \includegraphics[width=1.0\linewidth]{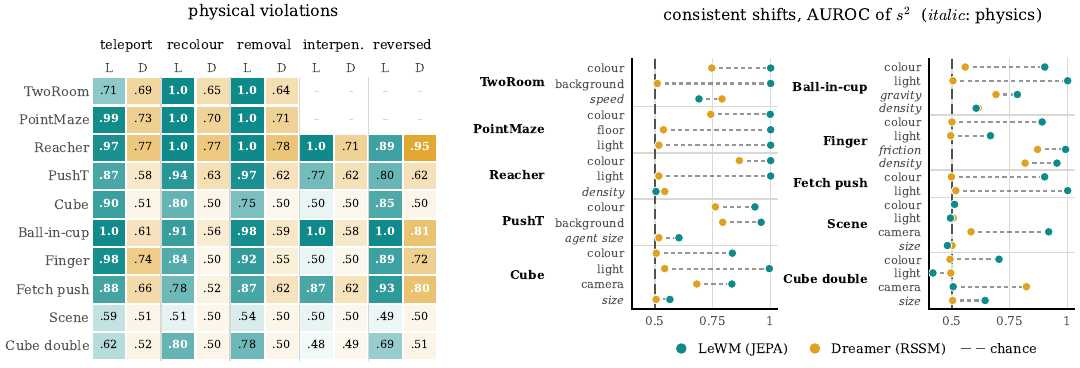}
\caption{\textbf{Physical violations and consistent shifts.} AUROC of $s^2$ for a violated frame (left) or a consistently altered environment (right) against clean windows. LeWM reports nearly every violation and shift; Dreamer remains near chance in many cases.}
   \label{fig:viol_heatmap}
\end{figure*}

Sections~\ref{sec:analyitcal_framework} and~\ref{sec:sensitivity_results} diagnosed an attenuated uncertainty sensitivity in Dreamer in contrast to LeWM. An uncertainty that responds to a state perturbation at a fraction of its size is not expected to serve tasks that rely on a well-structured uncertainty, such as detecting unfamiliar or physically inconsistent inputs or steering a planner. The two mechanisms sharpen this expectation: brief perturbations should be harder to register than persistent ones, and perturbations of entities that received little training signal should go unnoticed at any persistence. We now evaluate how far these expectations carry over to five uncertainty-dependent downstream tasks; exact protocols are given in Appendix B.

\paragraph{Unfamiliar-state detection.}
One quadrant of the body coordinates (covering about a quarter of the training data) of the state space is withheld when the Gram matrix $G$ of Eq.~\eqref{eq:laplace} is accumulated. Windows from the withheld quadrant are then scored against windows from the remaining data by the detection AUROC. Table~\ref{tab:detection_main} shows that LeWM separates the withheld region with AUROC $\ge 0.96$ in six of ten environments on the body and $\ge 0.97$ on the object in five of seven. Dreamer lies between $0.48$ and $0.98$ and is below LeWM in thirteen of seventeen cells. Where Dreamer leads, in Finger and on the body of Scene and Cube double, the withheld entity is the one holding the largest share of the reconstruction gradient, whereas on the small objects of the same two environments its AUROC falls to $0.67$ and $0.48$, consistent with the entity-share mechanism of Section~\ref{sec:sensitivity_results}.

\begin{figure}[!h]
  \centering
   \includegraphics[width=1.0\linewidth]{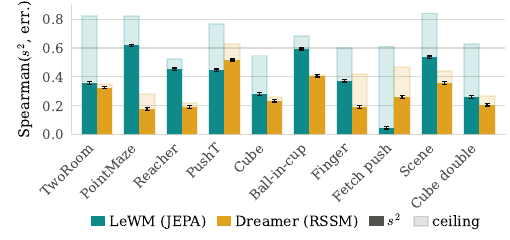}
\caption{\textbf{Error ranking.} Spearman correlation of $s^2$ with the model's own one-step error (solid) and the ceiling of a linear readout from the same features (pale). LeWM ranks errors better and its features hold further information; Dreamer's $s^2$ sits near a low ceiling.}
   \label{fig:error_ranking}
\end{figure}

\begin{figure*}[!h]
  \centering
   \includegraphics[width=1.0\linewidth]{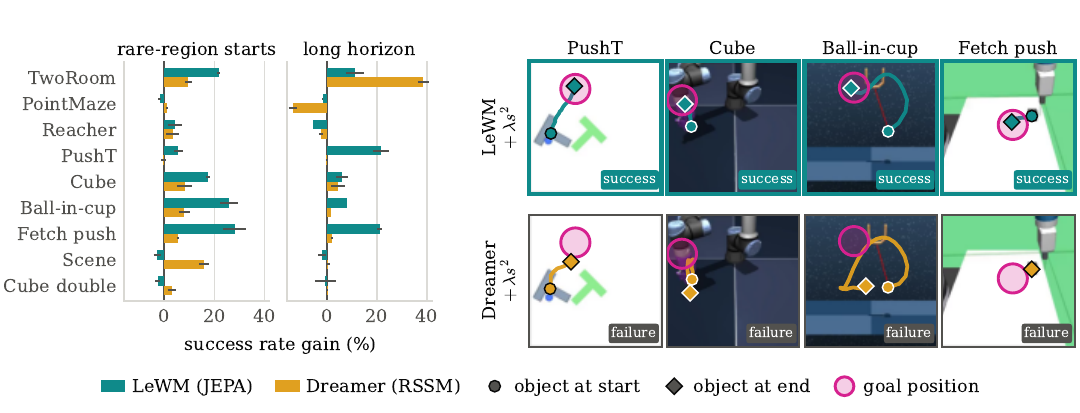}
\caption{\textbf{Uncertainty-penalised planning.} Left: change in success rate from adding the $s^2$ penalty for rare-region starts and long-horizon goals. Right: visual examples of planned trajectories for both models. Finger is omitted, as its spin task has no static goal state.}
   \label{fig:planning_combo}
\end{figure*}

\paragraph{Error ranking.}
On held-out transitions we compute each model's own one-step prediction error, the embedding residual $\|\hat{\mathbf z}_{t+1}-\mathbf z_{t+1}\|$ for LeWM and the posterior--prior KL for Dreamer, and report the Spearman correlation between $s^2$ and that error. As a ceiling we fit a ridge regression from $\boldsymbol\psi$ and $\log s^2$ to the error, the correlation an optimal linear readout of the same features attains.
Figure~\ref{fig:error_ranking} shows that LeWM ranks errors better in eight of ten environments, and that the two models differ in their ceilings. Dreamer's $s^2$ lies within $0.06$ of its ceiling in six environments, and the ceiling itself is low ($0.22$--$0.63$): the recurrent state carries little information about the coming error, and no readout of it would rank errors substantially better. LeWM's ceilings are higher ($0.52$--$0.84$) and exceed Dreamer's in every environment, so the features retain information about the coming error beyond what the Laplace head uses, leaving room for stronger estimators.

\paragraph{Physical-violation detection.}

The last frame of a held-out window is made inconsistent with the physics or the history of the window, by a teleport of the controlled entity, a recolouring or removal of the object, an interpenetration, or a reversal of the motion. Each violated window is scored against its clean counterpart.
Figure~\ref{fig:viol_heatmap} shows that LeWM detects every violation at $\ge 0.75$ in the eight single-object environments, except interpenetration in Cube and Finger, where the contact is not visible in the frame and both models are at chance. In the multi-object environments Scene and Cube double its AUROC drops to $0.49$--$0.80$. Dreamer exceeds $0.80$ only for the reversed motion in Reacher and Ball-in-cup and is at chance for every violation in Cube, Scene and Cube double, in which the reconstruction objective weights the object least (Section~\ref{sec:sensitivity_results}). Holding a violation over several context frames raises Dreamer's AUROC in part of the environments, as the persistence mechanism predicts, while LeWM is unaffected (Appendix C).

\paragraph{Response to consistent distribution shifts.}
In this downstream task the environment itself is consistently altered. The factors of variation of the stable-worldmodel suite are applied to held-out windows: \emph{visual} shifts re-render the recorded states under altered colour, background, lighting or camera, \emph{physics} shifts replay the recorded actions under altered speed, density, size, gravity or friction.
Figure~\ref{fig:viol_heatmap} shows that LeWM reports visual shifts with AUROC $\ge 0.83$ in 17 of 23 cases, whereas Dreamer exceeds $0.6$ in only seven and remains near chance for lighting and background throughout. Physics shifts are harder for both models, with AUROC between $0.48$ and $0.99$ and no consistent leader; Dreamer leads clearly only for the speed of TwoRoom and the camera of Cube double.

\paragraph{Uncertainty-penalised planning.}
Both models plan with CEM on the frozen predictor, with $s^2$ along the imagined trajectory added to the goal cost at weight $\lambda$. Two tasks lead the planner through sparsely covered states, rare-region starts and long-horizon goals; we report the change in success rate over 200 episodes. Figure~\ref{fig:planning_combo} shows gains for LeWM in six of nine environments under rare-region starts and five under the long horizon, by 17 to 28 points in TwoRoom, Cube, Ball-in-cup and Fetch push; Dreamer's gains stay mostly below 10 points, with TwoRoom and Scene as exceptions. PointMaze, Scene and Cube double gain nothing for either model. Appendix C includes baseline success rates. 

\paragraph{Summary.}
Across the five tasks, JEPA's non-attenuated uncertainty is the more useful signal: it separates withheld states, flags single-frame violations, registers shifted environments and translates into planning gains. Dreamer leads only where a change persists over its context or concerns the entity that dominates its reconstruction, in line with the two attenuation mechanisms. Appendix C confirms the mechanism vice versa: a Laplace head on Dreamer's posterior feature, read before the GRU update, detects violations and dynamics shifts at LeWM's level. The two paradigms shape their features differently, and uncertainty estimation inherits that difference.

\section{Discussion}
\label{sec:discussion}
We presented a wide comparison of post-hoc epistemic uncertainty between two dominant world-model paradigms (JEPA and RSSMs) across ten simulated visual-control environments. Tracing a controlled state perturbation through every stage showed how the paradigms shape the features an estimator reads, disclosed two sources of attenuation in Dreamer-like models, the recurrent update and the reconstruction objective, and showed that the difference carries over to five uncertainty-dependent tasks.
\textbf{Limitations.} The study contrasts one instance per paradigm, so the RSSM findings concern Dreamer-type models; it uses last-layer estimators only; the trace requires a simulator, so pretrained world models are not covered.
Future work should test the mechanisms causally, by removing the reconstruction term or altering the recurrent gate, extend the analysis to large pretrained JEPAs such as V-JEPA~2, and examine the relation between the physical sensitivity of JEPA representations and their uncertainty, which our violation results suggest is close.

\section{LLM USAGE STATEMENT}

Large language models (Claude Opus and Claude Fable, Anthropic) were used in three ways. First, as writing support: the authors determined the intellectual content of every section, including its claims, arguments and structure. The model's role was limited to wording: it proposed formulations and shortened existing text to fit the page limit, and the authors revised all such text. Second, for implementation support in writing and debugging the experiment code. Third, for producing the plotting scripts behind the figures. The research questions, the analytical framework, the experimental design, the selection and interpretation of results, and all scientific claims are the authors' own, and every generated text, code and figure was checked by the authors.

{
    \small
    \bibliographystyle{ieeenat_fullname}
    \bibliography{main}
}

\clearpage
\setcounter{page}{1}
\maketitlesupplementary

\section*{A Additonal Defintions}

\paragraph{Formal definition of the stage chain}
\label{app:chain}
Let $\mathcal{X}$ be the state space of the environment, $\mathcal{O}$ the observation space and $H$ the context length. A window at time $t$ is the pair $\mathbf{w}_t = (\mathbf{o}_{t-H+1:t}, \mathbf{a}_{t-H+1:t})$ with $\mathbf{o}_s = R(\mathbf{x}_s)$ for a renderer $R:\mathcal{X}\to\mathcal{O}$. The frozen model defines three maps,
\begin{align}
  E &: \mathcal{O} \to \mathbb{R}^{d_z}, \qquad \mathbf{z}_s = E(\mathbf{o}_s), \\
  \Phi &: \big(\mathbb{R}^{d_z}\big)^{H} \times \mathcal{A}^{H} \times \mathcal{H} \to \mathbb{R}^{F}, \notag \\
      &\qquad \boldsymbol{\psi}_t = \Phi\big(\mathbf{z}_{t-H+1:t}, \mathbf{a}_{t-H+1:t}, \mathbf{h}_{t-H+1}\big), \\
  U &: \mathbb{R}^{F} \to \mathbb{R}_{\ge 0}, \qquad u_t = U(\boldsymbol{\psi}_t;\, \mathcal{D}),
\end{align}
where $\mathcal{H}$ is the space of recurrent states and $\mathbf{h}_{t-H+1}$ the state carried into the window. For LeWM $\mathcal{H}$ is trivial and $\Phi$ is the feed-forward predictor acting on the $H$ encoded frames (Sec.~\ref{sec:chain}); for Dreamer $\Phi$ is the RSSM filter applied $H$ times, each step composing the posterior readout, the GRU update and the prior features (Sec.~\ref{sec:chain}), and its intermediate outputs $\mathbf{q}_s$ and $\mathbf{h}_{s+1}$ are stage outputs in the sense below. The chain of the main text,
\begin{equation}
  \mathbf{x} \xrightarrow{R} \mathbf{o} \xrightarrow{E} \mathbf{z}
  \xrightarrow{\Phi} \boldsymbol{\psi} \xrightarrow{U} u ,
\end{equation}
is the composition $u_t = U \circ \Phi \circ E^{\otimes H} \circ R^{\otimes H}$ applied to $\mathbf{x}_{t-H+1:t}$, with $F_\theta = \Phi \circ E^{\otimes H} \circ R^{\otimes H}$ the map from states to the predictor feature. A \emph{stage} is any map in this composition together with its output; a stage output is a deterministic function of the previous one, except for the sampling of the stochastic state from the posterior, which we evaluate at its mode when tracing. The relative displacement of Eq.~\eqref{eq:rho} is defined at every stage output, and the sensitivity $\kappa$ of Eq.~\eqref{eq:kappa} is its value at the last stage. Since $U$ depends on the window only through $\boldsymbol{\psi}_t$, a perturbation of any $\mathbf{x}_s$ in the window affects $u_t$ if it affects $\boldsymbol{\psi}_t$.

\paragraph{Ensemble.}
Following the common practice for world models~\cite{sekar2020planning,berger2026biased},
the ensemble consists of $M$ heads $f^{(1)},\dots,f^{(M)}$ that take the frozen
predictor feature as input and replace the final linear layer of the predictor.
Each head is an MLP with one hidden layer of width $512$,
$f^{(m)}:\mathbb{R}^{F}\to\mathbb{R}^{D}$, trained from its own initialisation
on a bootstrap resample of the training transitions with the original one-step
loss of the predictor; the world model itself stays frozen. We use $M=5$.
For a regression head (LeWM) the uncertainty is the disagreement of the
predictions,
\begin{align}
  u_{\mathrm{ens}}(\boldsymbol{\psi})
    &= \frac{1}{M}\sum_{m=1}^{M}
       \bigl\lVert f^{(m)}(\boldsymbol{\psi})-\bar f(\boldsymbol{\psi})\bigr\rVert^{2},
  \\
  \bar f(\boldsymbol{\psi})
    &= \frac{1}{M}\sum_{m=1}^{M} f^{(m)}(\boldsymbol{\psi}),
\end{align}
and for a categorical head (Dreamer) it is the Jensen--Shannon divergence of
the $M$ predicted distributions, averaged over the categorical groups.
Like $s^{2}$, $u_{\mathrm{ens}}$ depends on a test window only through
$\boldsymbol{\psi}$: the ensemble can register a perturbation of the environment state
only if $\boldsymbol{\psi}$ does. Appendix C display results for ensemble experiments.

\section*{B Implementation Details and Experimental Setting}
\label{app:dreamer}

Both models are trained following the protocol of LeWM, with AdamW, bf16 precision and a single checkpoint per environment that is then frozen. The post-hoc estimator is fitted on this frozen model. 

\paragraph{LeWM (JEPA).}
We use the LeWorldModel recipe~\citep{maes2026leworldmodel}. The encoder $E$ is a ViT-Tiny (patch size 14, $224\times224$ input, trained from scratch) whose class token, passed through a two-layer MLP projector with BatchNorm (hidden width 2048), gives the frame embedding $\mathbf z_t \in \mathbb R^{192}$. Actions are embedded to the same width by a linear embedder. The predictor $\Phi$ is a causal transformer over the last $H=3$ frames (depth 6, 16 heads of width 64, MLP width 2048, dropout 0.1, learned positional embedding) followed by a prediction MLP (hidden width 2048, BatchNorm, output 192). The predictor feature $\boldsymbol\psi_t \in \mathbb R^{2048}$ is the input of the final linear layer of this MLP and $W \in \mathbb R^{192\times 2048}$ its weight. Training minimises
\begin{equation}
  \mathcal L_{\mathrm{LeWM}} = \big\|\hat{\mathbf z}_{t+1} - \operatorname{sg}[\mathbf z_{t+1}]\big\|^2 + \lambda\,\mathcal R_{\mathrm{SIGReg}}(\mathbf z),
\end{equation}
$ \lambda = 0.09$, with $\mathcal R_{\mathrm{SIGReg}}$ the sketched isotropic-Gaussian regulariser of \citet{maes2026leworldmodel}: the Epps–Pulley statistic of the embeddings along 1024 random unit projections against a standard normal, evaluated on 17 knots in $[0,3]$, which prevents collapse without a target network or negatives. Batch size 128, learning rate $5\cdot10^{-5}$, weight decay $10^{-3}$, gradient clipping at 1.0. For LeWM we use the official checkpoints of \citet{maes2026leworldmodel} where available and train all other models for 100 epochs with early stopping on the validation loss.

\begin{figure*}[!h]
  \centering
   \includegraphics[width=1.0\linewidth]{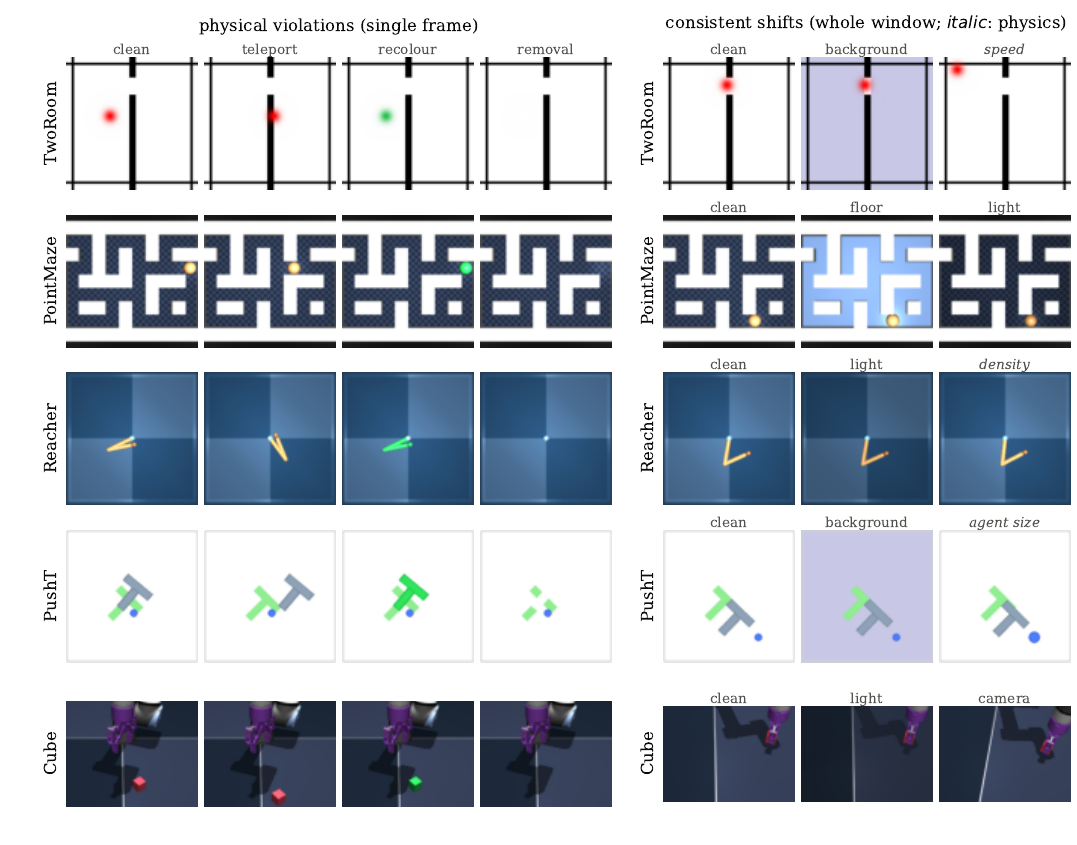}
\caption{\textbf{Examples of physical violations and consistent environment shifts.}}
   \label{fig:violations_examples}
\end{figure*}

\paragraph{DreamerV3 (RSSM).}
We use the DreamerV3 world model~\citep{hafner2025mastering} at its native $64\times64$ resolution in the ``small'' configuration: a four-stage CNN encoder (base depth 32, channel LayerNorm, SiLU) producing a 4096-d embedding $\mathbf z_t$; a layer-normed GRU with deterministic state $\mathbf h_t \in \mathbb R^{512}$; a stochastic state $\mathbf s_t$ of 32 categorical groups with 32 classes each ($1024$-d one-hot), with 1\% uniform mixing of the logits; posterior and prior heads as MLPs of width 512; a transposed-CNN decoder from the concatenated feature $[\mathbf h_t, \mathbf s_t]$; and the symlog two-hot reward head and the continuation head of the original recipe, which receive zero reward and constant continuation on our data and are inert. The predictor feature $\boldsymbol\psi_{t+1} \in \mathbb R^{512}$ is the hidden activation of the prior head before its final linear layer $W \in \mathbb R^{1024\times 512}$, which outputs the prior logits. Training minimises the DreamerV3 world-model objective \begin{align}
  \mathcal L_{\mathrm{Dreamer}} &= \tfrac12\,\big\|\mathbf o_t - \hat{\mathbf o}_t\big\|^2
  + \beta_{\mathrm{dyn}}\,\max\!\big(1, \mathrm{KL}[\operatorname{sg}(q_t)\,\|\,p_t]\big) \notag\\
  &\quad + \beta_{\mathrm{rep}}\,\max\!\big(1, \mathrm{KL}[q_t\,\|\,\operatorname{sg}(p_t)]\big),
\end{align}
with the squared error summed over pixels in $[-0.5, 0.5]$ scale, free bits of one nat on the KL summed over groups, and KL balancing $\beta_{\mathrm{dyn}} = 0.8$, $\beta_{\mathrm{rep}} = 0.2$. Sequences of 16 model steps, batch size 64 (1024 frames per update, the DreamerV3 token budget), learning rate $10^{-4}$, no weight decay, gradient clipping at 1000; again we train for 100 epochs with early stopping o validation loss.

\begin{figure*}[!h]
  \centering
   \includegraphics[width=1.0\linewidth]{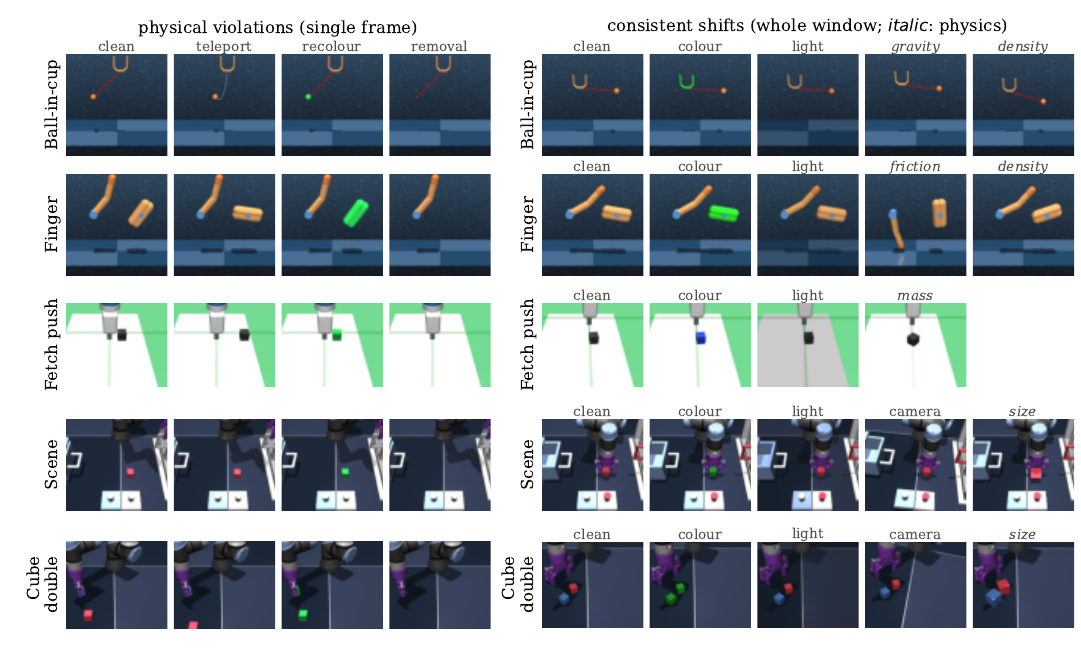}
\caption{\textbf{Examples of physical violations and consistent environment shifts.}}
   \label{fig:violations_examples_2}
\end{figure*}

\paragraph{Last-layer Laplace.}
For both models the Gram matrix $G$ of Eq.~\eqref{eq:laplace} is accumulated over the predictor features of teacher-forced one-step predictions on the training split. For LeWM the noise and prior variances $(\sigma^2, \tau^2)$ are selected on the validation split by the Gaussian negative log-likelihood of the held-out residuals, on a grid of 15 values of $\sigma^2$ around the maximum-likelihood residual variance and 17 values of $\tau^2$ in $[10^{-6}, 10^{2}]$. For Dreamer the head outputs categorical logits and no residual exists, so both hyperparameters are set in closed form from the training split. The noise variance is matched to the curvature of the softmax likelihood,
\begin{equation}
  \sigma^2 \;=\; \frac{C}{\bar t}, \qquad
  \bar t \;=\; \frac{1}{NK}\sum_{n=1}^{N}\sum_{k=1}^{K}\Big(1-\sum_{c=1}^{C} p_{n,k,c}^{2}\Big),
\end{equation}
the Gauss--Newton trace of one categorical group averaged over the $N$ training steps and the $K$ groups, with $p_{n,k,c}$ the prior probability of class $c$ in group $k$ at step $n$; the prior variance maximises the Laplace evidence
\begin{equation}
  \tau^2 = \arg\max_{\tau^2}\;-\tfrac{1}{2}\,\frac{\|\mathbf{W}\|^2}{\tau^2} \;-\; \tfrac{1}{2}\,KC \sum_{i}\log\!\Big(1+\frac{\tau^2\lambda_i}{\sigma^2}\Big),
\end{equation}
where $\lambda_i$ are the eigenvalues of the Gram matrix, $\mathbf{W}$ the head's weights and $K\times C$ its output dimension. Ensemble heads (Appendix A) are trained likewise.

\subsection*{Reconstruction gradient of an entity}
\label{app:mech}

\begin{figure*}[!h]
\centering
\includegraphics[width=\textwidth]{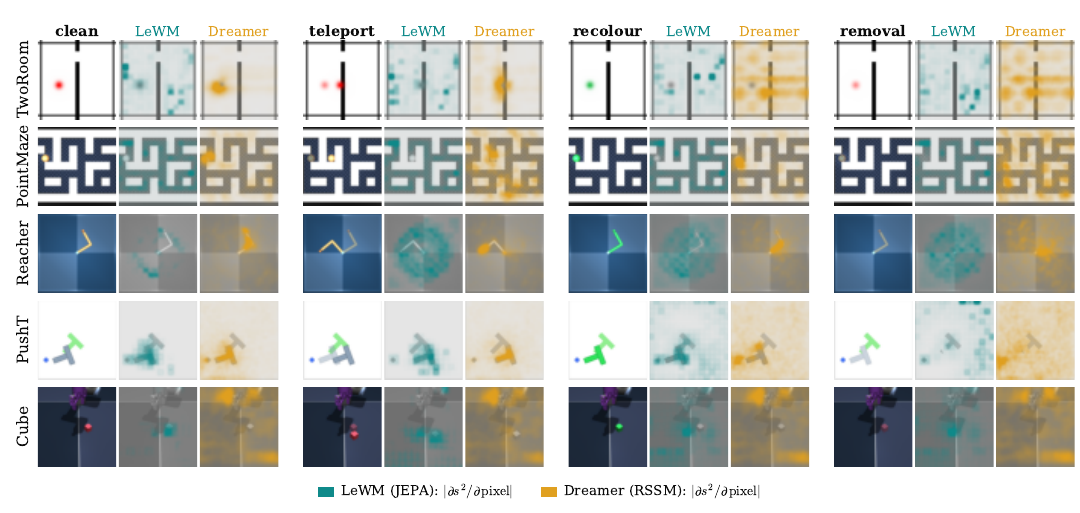}
\caption{\textbf{Pixel sensitivity of the uncertainty, part 1.} SmoothGrad map $S(p)$ of Eq.~\eqref{eq:smoothgrad} for the clean frame and three violations (frame, LeWM map, Dreamer map), one window per environment.}
\label{fig:maps_a}
\end{figure*}

\paragraph{Entity (body and object) masks and pixel area.}
For a frame $\mathbf o$ rendered at state $\mathbf x$, the pixel mask $M_e \subset \{1,\dots,P\}$ of an entity $e \in \{\mathrm b, \mathrm o\}$, the body or the object, is obtained in the simulator: the geoms of the entity are hidden, the frame is re-rendered from the same state, and $M_e$ is the set of pixels at which the two renders differ, downsampled to the $64\times64$ resolution the decoder outputs. The body mask excludes object pixels, $M_{\mathrm b} \leftarrow M_{\mathrm b}\setminus M_{\mathrm o}$, and the remaining pixels form the background mask $M_{\mathrm r}$ ($e=\mathrm r$).

\paragraph{Gradient of the reconstruction term on an entity.}
Let $\mathbf f_t = [\mathbf h_t, \mathbf s_t]$ be the feature the decoder reads and $\hat{\mathbf o}_t = D(\mathbf f_t)$ its reconstruction. The reconstruction term of $\mathcal L_{\mathrm{RSSM}}$, a Gaussian likelihood with unit variance, restricted to the pixels of entity $e$, is
\begin{equation}
  \ell_e(\mathbf f_t) \;=\; \tfrac12 \sum_{p \in M_e} \big\| \mathbf o_{t,p} - \hat{\mathbf o}_{t,p} \big\|^2 ,
\end{equation}
with the sum over colour channels inside the norm. We define the reconstruction gradient of the entity as
\begin{equation}
  G_e \;=\; \big\| \nabla_{\mathbf f_t}\, \ell_e(\mathbf f_t) \big\| ,
  \label{eq:rec_entity}
\end{equation}
the norm of the pull that the pixels of $e$ exert on the latent state. It is evaluated at the posterior state of held-out windows: the context is filtered with the posterior, $\mathbf f_t$ at the last frame is detached and treated as the variable, the frame is decoded, and $\ell_e$ is differentiated with respect to $\mathbf f_t$ for $e \in \{\mathrm b, \mathrm o, \mathrm r\}$ separately. $G_e$ and the ratio $G_{\mathrm o}/G_{\mathrm b}$ are reported as medians over $N = 150$ windows per environment. The entity's share of the reconstruction gradient,
\begin{equation}
  s_e \;=\; \frac{G_e}{G_{\mathrm b} + G_{\mathrm o}} ,
  \label{eq:pull}
\end{equation}
is the regressor of Appendix~\ref{app:regression}; body-only environments have $s_{\mathrm b} = 1$.

\paragraph{Mechanistic Influence Factors in Section~\ref{sec:sensitivity_results}}
\begin{figure*}[h]
\centering
\includegraphics[width=\textwidth]{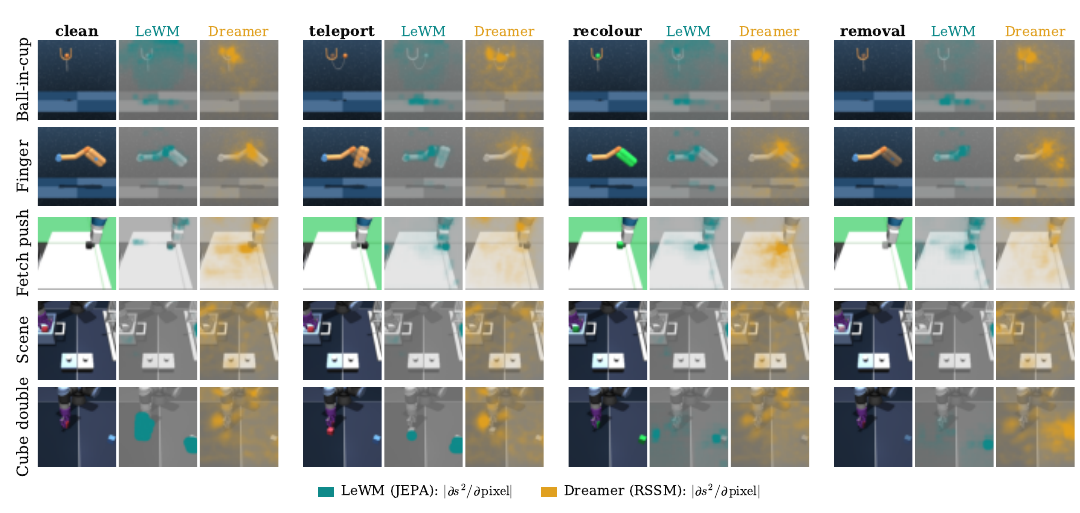}
\caption{\textbf{Pixel sensitivity of the uncertainty, part 2.} As Fig.~\ref{fig:maps_a} for Ball-in-cup, Finger, Fetch push, Scene and Cube double.}
\label{fig:maps_b}
\end{figure*}

Figure~\ref{fig:mech-factors}c plots $G_e$ against $|M_e|/P$ with one point per entity (that is, body or object) and environment; the Spearman correlation is over these points. Figure~\ref{fig:mech-factors}d plots $\kappa^{\mathrm o}/\kappa^{\mathrm b}$ at $K = 1$, the object-to-body ratio of the sensitivity of Eq.~\eqref{eq:kappa}, against $G_{\mathrm o}/G_{\mathrm b}$, one point per object environment and model; the Spearman correlations are over  environments, separately for Dreamer and LeWM. $G_e$ is a quantity only derivable for Dreamer, since LeWM has no decoder; the LeWM points use the same abscissa and serve as the control in this panel.

\subsection*{Two-factor regression of sensitivity}
\label{app:regression}

\paragraph{Regression Analysis.}
Each observation of the regression is one environment, one entity (body, or object where the environment has one) and one persistence $K$; its response is $\log\kappa$, the sensitivity of the uncertainty head to a one-standard-deviation displacement of the entity's coordinates held over the last $K$ context frames (Sec.~\ref{sec:trace}), taken as the median over the traced windows. Dreamer observations use $K=1,\dots,8$, its full context: 17 environment--entity pairs $\times$ 8 values of $K$ $=$ 136 observations. LeWM sees three frames, so a displacement held over $K>3$ frames is the same input as at $K=3$; its observations are restricted to $K\le 3$. For each world model separately we fit
\begin{equation}
\log\kappa \;=\; \beta_0 \;+\; \beta_1 \log s_e \;+\; \beta_2 \log\!\bigl(1-(1-g)^K\bigr) \;+\; \varepsilon .
\label{eq:twofactor}
\end{equation}
The first regressor is the objective factor. For Dreamer, $s_e = G_e/(G_b+G_o)$ is the entity's share of the reconstruction-gradient norm on the latent (previous paragraph); body-only environments have $s_b=1$. LeWM has no reconstruction loss and hence no gradient share; its regressor is the model-free proxy, the entity's share of the image, $s_e = |M_e|/(|M_b|+|M_o|)$. The second regressor is the architecture factor: the fraction of a displacement that a leaky recurrent state with gate $g$ retains after being refreshed for $K$ steps. The gate is fitted once on Dreamer's GRU-stage displacements $\rho_{\mathbf h}(K)$ by non-linear least squares with a free intercept per environment--entity pair ($\hat g = 0.25$, Fig.~\ref{fig:mech-factors}a) and held fixed in Eq.~\eqref{eq:twofactor}. For LeWM the term is a control: it has no recurrent state, so $\beta_2$ should be zero. Cells of one environment share an encoder, a dataset and a policy, so their residuals are correlated. Standard errors are cluster-robust by environment (sandwich estimator with the small-sample factor $\tfrac{G}{G-1}\tfrac{n-1}{n-k}$, $G=10$ clusters), and $p$-values and 95\% intervals use a $t$ distribution with $G-1=9$ degrees of freedom.

\paragraph{Fitted Regression Equations.}
\begin{align}
\text{Dreamer:}\quad \log\kappa &= 0.25 + 1.21\,\log s_e \notag\\
  &\quad + 0.87\,\log\!\bigl(1-(1-g)^K\bigr), & R^2 &= 0.65, \\
\text{LeWM:}\quad \log\kappa &= 1.09 + 0.40\,\log s_e \notag\\
  &\quad - 0.01\,\log\!\bigl(1-(1-g)^K\bigr), & R^2 &= 0.11,
\end{align}
with 95\% intervals $\pm 0.65$ ($p=0.002$) and $\pm 0.30$ ($p<0.001$) for Dreamer's two slopes, and $\pm 0.37$ ($p=0.04$) and $\pm 0.03$ ($p=0.40$) for LeWM's. The intercepts are $\log\kappa$ for a body-only entity at full persistence: $\kappa\approx 1.3$ for Dreamer and $\approx 3$ for LeWM, matching Fig.~\ref{fig:kappa}. 

\subsection*{Downstream task protocols}
\label{app:tasks}

\begin{figure*}[!h]
\centering
\includegraphics[width=\textwidth]{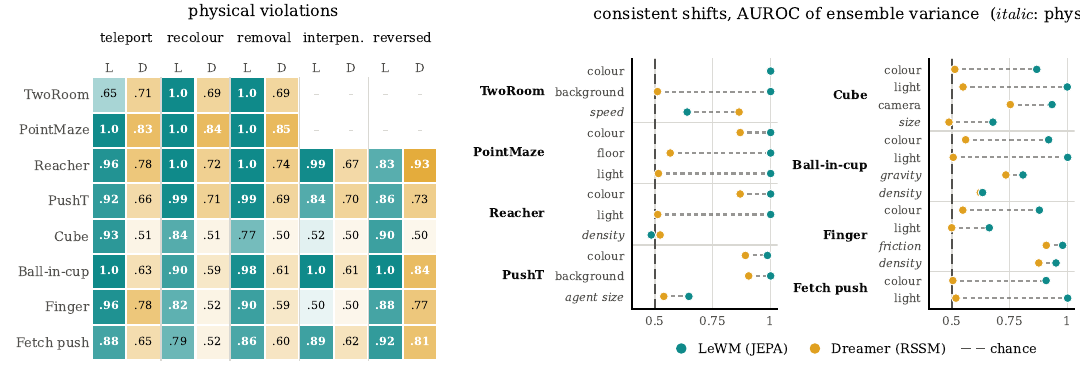}
\caption{\textbf{Physical violations and consistent shifts, post-hoc ensemble.} As Fig.~\ref{fig:viol_heatmap}, with the ensemble disagreement in place of $s^2$.}
\label{fig:viol_heatmap_ens}
\end{figure*}

All tasks use, if not stated otherwise, the frozen models and the Laplace head of Section ~\ref{sec:analyitcal_framework}, held-out episodes, and frames rendered by the simulator from the recorded states so that both models see the same renderer. $u = s^2$ throughout. AUROC is the area under the ROC curve of $u$ for the positive class named in each task.

\paragraph{Unfamiliar-state detection.}
Let $(x_1, x_2)$ be the two position coordinates of a group (body: agent, arm or cup; object: block, cube or ball), and let $m_1, m_2$ be their medians over the training frames. The four quadrants $C_{\pm\pm} = \{\mathbf x : \pm(x_1 - m_1) > 0,\ \pm(x_2 - m_2) > 0\}$ each hold about a quarter of the training data. For each quadrant $C$ the Gram matrix of Eq.~\eqref{eq:laplace} is re-accumulated over the training features whose state lies outside $C$, with $(\sigma^2, \tau^2)$ unchanged, and the model itself is not modified. Held-out frames are then scored by $u$, the positive class being $\mathbf x \in C$. We report mean and standard deviation over the four quadrants; a size-matched random deletion of training frames gives the chance reference. For Finger, whose spinner has a single hinge angle, the quadrant is replaced by a half of the circle.

\paragraph{Error ranking.}
On held-out transitions each model's own one-step error is $\epsilon_t = \|\hat{\mathbf z}_{t+1} - \mathbf z_{t+1}\|^2$ for LeWM and the posterior–prior divergence $\mathrm{KL}[q_{t+1}\,\|\,p_{t+1}]$ for Dreamer. The score is the Spearman rank correlation $\rho_{\mathrm S}(u_t, \epsilon_t)$. The ceiling is a ridge regression from $[\boldsymbol\psi_t, \log u_t]$ to $\log \epsilon_t$, fitted on the training split and scored by the same rank correlation on the held-out split; it contains $u$ as a regressor, so ceiling $\ge$ score by construction, and measures how much of the coming error any linear readout of the same features can rank. Standard deviations are a block bootstrap over consecutive validation windows (blocks of 100, 500 resamples).

\paragraph{Physical-violation detection.}
We draw $N = 60$ held-out base states per environment, each with its eight-frame context window. A violation alters the last $K$ of these frames, and the altered window is scored against the unaltered one, the violated window being the positive class. The violations are defined on the simulator state or appearance and re-rendered: \emph{teleport}, the coordinates of the controlled entity displaced by one training standard deviation; \emph{recolour}, the object's material replaced; \emph{removal}, the object's geoms hidden; \emph{interpenetration}, the object sunk into its support or into the body (cube half into the table, ball into the cup wall, block onto the agent, puck under the gripper, Reacher wrist folded past its joint limit); \emph{reversed}, the object poses of the context taken in reverse order with velocities flipped, body untouched. Figure~\ref{fig:viol_heatmap} uses $K = 1$ for the first four and $K = 8$ for the reversed motion, which needs the context to be defined; Table~\ref{tab:persistence} holds the violations over $K \in \{1, 3, 8\}$. Standard deviations are a paired bootstrap over base states. Dreamer's $u$ is averaged over eight samples of the stochastic state. Figures \ref{fig:violations_examples} and \ref{fig:violations_examples_2} shows some visual examples of physical violations.

\paragraph{Consistent distribution shifts.}
We draw $N = 60$ held-out windows per environment, each of eight context frames and the following frame, and re-create every window under an altered environment, using the variation space of the stable-worldmodel suite. A \emph{visual} shift (colour, lighting, background, floor, camera) re-renders the recorded states under the altered variation, so frames change and dynamics do not. A \emph{physics} shift (speed, density, size, gravity, friction) replays the recorded actions from the recorded start state in the altered simulator, so the dynamics change; a \emph{replay} control does the same in the unaltered simulator and separates replay drift from the shift, giving AUROC $0.50$ throughout. The score is the AUROC of $u$ for shifted against clean windows. The camera shift is $8^\circ$ except in Scene, whose variation space bounds it at $5^\circ$, where it is $4^\circ$. Figures \ref{fig:violations_examples} and \ref{fig:violations_examples_2} shows some visual examples of physical violations.

\paragraph{Uncertainty-penalised planning.}
Both models plan actions with CEM (300 samples, 30 iterations, 30 elites, receding horizon of one action block of five simulator steps). For a candidate action sequence $\mathbf a$ with goal cost $C_{\mathrm{goal}}(\mathbf a)$, the penalised objective is
\begin{equation}
  C_\lambda(\mathbf a) \;=\; C_{\mathrm{goal}}(\mathbf a) \;+\; \lambda\,\frac{\operatorname{med}_{\mathbf a}[C_{\mathrm{goal}}]}{\operatorname{med}_{\mathbf a}[s^2]}\; s^2(\mathbf a),
  \label{eq:penalty}
\end{equation}
where $s^2(\mathbf a)$ is the Laplace variance of the predictor feature at the end of the imagined trajectory and the medians are over the candidates of the current solve, so that $\lambda$ is the weight of the penalty relative to the goal-cost scale and means the same for both model families. The goal cost is the embedding distance to the goal frame for LeWM and, for Dreamer, the decoder's pixel error to the goal frame (the posterior divergence in PushT). We report the change in success rate at the best $\lambda \in \{0.1, 0.3, 1\}$ against $\lambda = 0$ on the same episodes. \emph{Rare-region starts} draw the start state from the 2\% of training rows with the lowest occupancy on a grid over the group coordinates, which lie 15 to 50 times below the median row density; \emph{long horizon} triples the goal offset and the planning budget (75 and 150 simulator steps against 25 and 50). Each cell is 200 episodes per seed over three seeds; Finger is omitted since its spin task has no static goal state.

\section*{C Additional Results}
\label{sec:appendix_addtional_results}

\begin{table*}[h]
\centering
\footnotesize\setlength{\tabcolsep}{2.5pt}
\caption{\textbf{Data of Fig.~\ref{fig:trace_analysis}.} Relative displacement per one-std perturbation held over the last $K$ context frames: Dreamer's posterior (post.), recurrent state $\mathbf{h}_{t+1}$ and predictor feature $\boldsymbol{\psi}_{t+1}$, and LeWM's predictor feature $\boldsymbol{\phi}_t$. Mean over the perturbed coordinates $\pm$ sd over coordinates.}
\label{tab:trace}
\begin{tabular}{lcccccccccccc}
\toprule
& \multicolumn{4}{c}{$K=1$} & \multicolumn{4}{c}{$K=3$} & \multicolumn{4}{c}{$K=8$} \\ \cmidrule(lr){2-5} \cmidrule(lr){6-9} \cmidrule(lr){10-13}
Environment & post. & $\mathbf{h}_{t+1}$ & $\boldsymbol{\psi}_{t+1}$ & $\boldsymbol{\phi}_t$ & post. & $\mathbf{h}_{t+1}$ & $\boldsymbol{\psi}_{t+1}$ & $\boldsymbol{\phi}_t$ & post. & $\mathbf{h}_{t+1}$ & $\boldsymbol{\psi}_{t+1}$ & $\boldsymbol{\phi}_t$ \\
\midrule
TwoRoom & 1.27{\scriptsize$\pm$.06} & 0.29{\scriptsize$\pm$.04} & 0.28{\scriptsize$\pm$.07} & 1.19{\scriptsize$\pm$.01} & 1.29{\scriptsize$\pm$.05} & 0.49{\scriptsize$\pm$.08} & 0.52{\scriptsize$\pm$.14} & 1.20{\scriptsize$\pm$.00} & 1.11{\scriptsize$\pm$.08} & 0.94{\scriptsize$\pm$.10} & 1.03{\scriptsize$\pm$.10} & 1.20{\scriptsize$\pm$.00} \\
PointMaze & 1.05{\scriptsize$\pm$.01} & 0.28{\scriptsize$\pm$.00} & 0.21{\scriptsize$\pm$.00} & 1.05{\scriptsize$\pm$.05} & 1.02{\scriptsize$\pm$.02} & 0.54{\scriptsize$\pm$.00} & 0.46{\scriptsize$\pm$.01} & 1.09{\scriptsize$\pm$.05} & 0.90{\scriptsize$\pm$.02} & 1.10{\scriptsize$\pm$.05} & 1.13{\scriptsize$\pm$.05} & 1.09{\scriptsize$\pm$.05} \\
Reacher & 1.22{\scriptsize$\pm$.12} & 0.16{\scriptsize$\pm$.02} & 0.11{\scriptsize$\pm$.01} & 1.00{\scriptsize$\pm$.01} & 1.21{\scriptsize$\pm$.13} & 0.24{\scriptsize$\pm$.00} & 0.18{\scriptsize$\pm$.01} & 1.08{\scriptsize$\pm$.06} & 1.14{\scriptsize$\pm$.27} & 0.73{\scriptsize$\pm$.38} & 0.77{\scriptsize$\pm$.47} & 1.08{\scriptsize$\pm$.06} \\
PushT & 1.19{\scriptsize$\pm$.16} & 0.27{\scriptsize$\pm$.07} & 0.30{\scriptsize$\pm$.08} & 1.01{\scriptsize$\pm$.21} & 1.02{\scriptsize$\pm$.13} & 0.60{\scriptsize$\pm$.17} & 0.65{\scriptsize$\pm$.19} & 1.01{\scriptsize$\pm$.20} & 0.86{\scriptsize$\pm$.09} & 0.96{\scriptsize$\pm$.16} & 0.94{\scriptsize$\pm$.14} & 1.01{\scriptsize$\pm$.20} \\
Cube & 0.94{\scriptsize$\pm$.60} & 0.10{\scriptsize$\pm$.07} & 0.10{\scriptsize$\pm$.08} & 0.72{\scriptsize$\pm$.13} & 0.92{\scriptsize$\pm$.61} & 0.17{\scriptsize$\pm$.13} & 0.17{\scriptsize$\pm$.13} & 0.73{\scriptsize$\pm$.13} & 0.85{\scriptsize$\pm$.54} & 0.40{\scriptsize$\pm$.36} & 0.43{\scriptsize$\pm$.40} & 0.73{\scriptsize$\pm$.13} \\
Ball-in-cup & 1.10{\scriptsize$\pm$.09} & 0.15{\scriptsize$\pm$.01} & 0.12{\scriptsize$\pm$.01} & 1.00{\scriptsize$\pm$.07} & 1.05{\scriptsize$\pm$.12} & 0.27{\scriptsize$\pm$.03} & 0.22{\scriptsize$\pm$.03} & 0.95{\scriptsize$\pm$.10} & 1.02{\scriptsize$\pm$.13} & 0.42{\scriptsize$\pm$.10} & 0.37{\scriptsize$\pm$.11} & 0.95{\scriptsize$\pm$.10} \\
Finger & 0.95{\scriptsize$\pm$.06} & 0.27{\scriptsize$\pm$.07} & 0.28{\scriptsize$\pm$.07} & 1.07{\scriptsize$\pm$.27} & 0.99{\scriptsize$\pm$.09} & 0.37{\scriptsize$\pm$.11} & 0.37{\scriptsize$\pm$.12} & 1.08{\scriptsize$\pm$.28} & 0.99{\scriptsize$\pm$.09} & 0.40{\scriptsize$\pm$.10} & 0.40{\scriptsize$\pm$.11} & 1.08{\scriptsize$\pm$.28} \\
Fetch push & 1.27{\scriptsize$\pm$.23} & 0.14{\scriptsize$\pm$.04} & 0.14{\scriptsize$\pm$.04} & 0.77{\scriptsize$\pm$.29} & 1.18{\scriptsize$\pm$.31} & 0.27{\scriptsize$\pm$.08} & 0.28{\scriptsize$\pm$.10} & 0.79{\scriptsize$\pm$.30} & 1.05{\scriptsize$\pm$.34} & 0.54{\scriptsize$\pm$.21} & 0.59{\scriptsize$\pm$.24} & 0.79{\scriptsize$\pm$.30} \\
Scene & 0.80{\scriptsize$\pm$.47} & 0.06{\scriptsize$\pm$.04} & 0.06{\scriptsize$\pm$.05} & 0.33{\scriptsize$\pm$.37} & 0.79{\scriptsize$\pm$.48} & 0.11{\scriptsize$\pm$.06} & 0.11{\scriptsize$\pm$.08} & 0.33{\scriptsize$\pm$.37} & 0.68{\scriptsize$\pm$.52} & 0.30{\scriptsize$\pm$.19} & 0.28{\scriptsize$\pm$.19} & 0.33{\scriptsize$\pm$.37} \\
Cube double & 0.98{\scriptsize$\pm$.39} & 0.06{\scriptsize$\pm$.03} & 0.06{\scriptsize$\pm$.02} & 0.57{\scriptsize$\pm$.32} & 0.97{\scriptsize$\pm$.40} & 0.12{\scriptsize$\pm$.04} & 0.12{\scriptsize$\pm$.04} & 0.57{\scriptsize$\pm$.32} & 0.84{\scriptsize$\pm$.46} & 0.40{\scriptsize$\pm$.22} & 0.42{\scriptsize$\pm$.24} & 0.57{\scriptsize$\pm$.32} \\
\bottomrule
\end{tabular}
\end{table*}

\paragraph{Stage trace.} Table~\ref{tab:trace} lists the stage displacements of Fig.~\ref{fig:trace_analysis} for all ten environments, including Scene and Cube double, which the figure omits. Each entry is the relative displacement of a representation per one-std perturbation of a state coordinate held over the last $K\in\{1,3,8\}$ context frames, where $1$ means the representation moves as much as the input; the mean is over the perturbed coordinates of body and object, the sd over coordinates. For Dreamer we report the posterior over the perturbed frame, the recurrent state $\mathbf{h}_{t+1}$ after the GRU update and the predictor feature $\boldsymbol{\psi}_{t+1}$ the Laplace head reads; for LeWM the predictor feature $\boldsymbol{\phi}_t$.

\subsection*{Physical Violations and Consistent Shifts}
\label{app:tables}

This section complements Fig.~\ref{fig:viol_heatmap}. Table~\ref{tab:persistence} holds the physical violations over increasing persistence, the violation applied to the last $K \in \{1, 3, 8\}$ context frames, so that $K=3$ covers LeWM's whole context and $K=8$ Dreamer's; Tables~\ref{tab:violations} and \ref{tab:shift} list the AUROC of $s^2$ with its bootstrap standard deviation for physical violations and every consistent shift experiments.

\begin{table*}[!h]
\centering
\small
\caption{\textbf{Violations held over $K$ frames.} AUROC of $s^2$ for a violated window against the clean window (mean of teleport, recolour and removal; object where present, otherwise the body), with the violation held in the last $K$ context frames. $K=3$ is LeWM's whole context, $K=8$ Dreamer's. Bold marks the better model. Error bars are evaluation variance for the fixed trained models (bootstrap over base states, clean and violated paired; sd of the mean over the three violation kinds).}
\label{tab:persistence}
\begin{tabular}{lcccccc}
\toprule
& \multicolumn{2}{c}{$K=1$} & \multicolumn{2}{c}{$K=3$} & \multicolumn{2}{c}{$K=8$} \\ \cmidrule(lr){2-3} \cmidrule(lr){4-5} \cmidrule(lr){6-7}
Environment & LeWM & Dreamer & LeWM & Dreamer & LeWM & Dreamer \\
\midrule
TwoRoom & \textbf{0.90}{\scriptsize$\pm$0.01} & 0.66{\scriptsize$\pm$0.01} & \textbf{0.87}{\scriptsize$\pm$0.01} & 0.74{\scriptsize$\pm$0.01} & \textbf{0.87}{\scriptsize$\pm$0.01} & 0.70{\scriptsize$\pm$0.02} \\
PointMaze & \textbf{1.00}{\scriptsize$\pm$0.00} & 0.71{\scriptsize$\pm$0.01} & \textbf{0.99}{\scriptsize$\pm$0.00} & 0.89{\scriptsize$\pm$0.01} & \textbf{0.99}{\scriptsize$\pm$0.00} & 0.98{\scriptsize$\pm$0.01} \\
Reacher & \textbf{0.99}{\scriptsize$\pm$0.00} & 0.77{\scriptsize$\pm$0.01} & \textbf{0.97}{\scriptsize$\pm$0.01} & 0.90{\scriptsize$\pm$0.01} & \textbf{0.97}{\scriptsize$\pm$0.01} & 0.96{\scriptsize$\pm$0.01} \\
PushT & \textbf{0.93}{\scriptsize$\pm$0.01} & 0.61{\scriptsize$\pm$0.01} & \textbf{0.91}{\scriptsize$\pm$0.01} & 0.76{\scriptsize$\pm$0.01} & \textbf{0.91}{\scriptsize$\pm$0.01} & 0.72{\scriptsize$\pm$0.02} \\
Cube & \textbf{0.82}{\scriptsize$\pm$0.01} & 0.50{\scriptsize$\pm$0.00} & \textbf{0.81}{\scriptsize$\pm$0.01} & 0.50{\scriptsize$\pm$0.00} & \textbf{0.81}{\scriptsize$\pm$0.01} & 0.50{\scriptsize$\pm$0.00} \\
Ball-in-cup & \textbf{0.96}{\scriptsize$\pm$0.01} & 0.59{\scriptsize$\pm$0.01} & \textbf{0.96}{\scriptsize$\pm$0.01} & 0.68{\scriptsize$\pm$0.01} & \textbf{0.96}{\scriptsize$\pm$0.01} & 0.78{\scriptsize$\pm$0.01} \\
Finger & \textbf{0.91}{\scriptsize$\pm$0.01} & 0.60{\scriptsize$\pm$0.01} & \textbf{0.92}{\scriptsize$\pm$0.01} & 0.62{\scriptsize$\pm$0.01} & \textbf{0.92}{\scriptsize$\pm$0.01} & 0.62{\scriptsize$\pm$0.01} \\
Fetch push & \textbf{0.84}{\scriptsize$\pm$0.01} & 0.60{\scriptsize$\pm$0.01} & \textbf{0.85}{\scriptsize$\pm$0.01} & 0.71{\scriptsize$\pm$0.01} & \textbf{0.85}{\scriptsize$\pm$0.01} & 0.78{\scriptsize$\pm$0.01} \\
\bottomrule
\end{tabular}
\end{table*}

\begin{table*}[!h]
\centering
\footnotesize\setlength{\tabcolsep}{3pt}
\caption{\textbf{Data of Fig.~\ref{fig:viol_heatmap}, violations.} AUROC of $s^2$ for a violated against the clean window, $\pm$ bootstrap sd over base states (clean and violated paired); L = LeWM, D = Dreamer; bold = better model; \mbox{--} = not applicable.}
\label{tab:violations}
\begin{tabular}{lcccccccccc}
\toprule
& \multicolumn{2}{c}{teleport} & \multicolumn{2}{c}{recolour} & \multicolumn{2}{c}{removal} & \multicolumn{2}{c}{interpen.} & \multicolumn{2}{c}{reversed} \\ \cmidrule(lr){2-3} \cmidrule(lr){4-5} \cmidrule(lr){6-7} \cmidrule(lr){8-9} \cmidrule(lr){10-11}
Environment & L & D & L & D & L & D & L & D & L & D \\
\midrule
TwoRoom & \textbf{0.71}{\scriptsize$\pm$.03} & 0.69{\scriptsize$\pm$.02} & \textbf{1.00}{\scriptsize$\pm$.00} & 0.65{\scriptsize$\pm$.02} & \textbf{1.00}{\scriptsize$\pm$.00} & 0.64{\scriptsize$\pm$.02} & -- & -- & -- & -- \\
PointMaze & \textbf{0.99}{\scriptsize$\pm$.00} & 0.73{\scriptsize$\pm$.02} & \textbf{1.00}{\scriptsize$\pm$.00} & 0.70{\scriptsize$\pm$.02} & \textbf{1.00}{\scriptsize$\pm$.00} & 0.71{\scriptsize$\pm$.02} & -- & -- & -- & -- \\
Reacher & \textbf{0.97}{\scriptsize$\pm$.01} & 0.77{\scriptsize$\pm$.02} & \textbf{1.00}{\scriptsize$\pm$.00} & 0.77{\scriptsize$\pm$.02} & \textbf{1.00}{\scriptsize$\pm$.00} & 0.78{\scriptsize$\pm$.02} & \textbf{1.00}{\scriptsize$\pm$.00} & 0.71{\scriptsize$\pm$.02} & 0.89{\scriptsize$\pm$.03} & \textbf{0.95}{\scriptsize$\pm$.02} \\
PushT & \textbf{0.87}{\scriptsize$\pm$.03} & 0.58{\scriptsize$\pm$.01} & \textbf{0.94}{\scriptsize$\pm$.02} & 0.63{\scriptsize$\pm$.02} & \textbf{0.97}{\scriptsize$\pm$.01} & 0.62{\scriptsize$\pm$.01} & \textbf{0.77}{\scriptsize$\pm$.03} & 0.62{\scriptsize$\pm$.02} & \textbf{0.80}{\scriptsize$\pm$.03} & 0.62{\scriptsize$\pm$.03} \\
Cube & \textbf{0.90}{\scriptsize$\pm$.02} & 0.51{\scriptsize$\pm$.01} & \textbf{0.80}{\scriptsize$\pm$.03} & 0.50{\scriptsize$\pm$.00} & \textbf{0.75}{\scriptsize$\pm$.03} & 0.50{\scriptsize$\pm$.00} & \textbf{0.50}{\scriptsize$\pm$.00} & 0.50{\scriptsize$\pm$.00} & \textbf{0.85}{\scriptsize$\pm$.02} & 0.50{\scriptsize$\pm$.00} \\
Ball-in-cup & \textbf{1.00}{\scriptsize$\pm$.00} & 0.61{\scriptsize$\pm$.02} & \textbf{0.91}{\scriptsize$\pm$.02} & 0.56{\scriptsize$\pm$.01} & \textbf{0.98}{\scriptsize$\pm$.01} & 0.59{\scriptsize$\pm$.01} & \textbf{1.00}{\scriptsize$\pm$.00} & 0.58{\scriptsize$\pm$.01} & \textbf{1.00}{\scriptsize$\pm$.00} & 0.81{\scriptsize$\pm$.02} \\
Finger & \textbf{0.98}{\scriptsize$\pm$.01} & 0.74{\scriptsize$\pm$.03} & \textbf{0.84}{\scriptsize$\pm$.02} & 0.50{\scriptsize$\pm$.01} & \textbf{0.92}{\scriptsize$\pm$.02} & 0.55{\scriptsize$\pm$.01} & \textbf{0.50}{\scriptsize$\pm$.00} & 0.50{\scriptsize$\pm$.00} & \textbf{0.89}{\scriptsize$\pm$.02} & 0.72{\scriptsize$\pm$.02} \\
Fetch push & \textbf{0.88}{\scriptsize$\pm$.02} & 0.66{\scriptsize$\pm$.02} & \textbf{0.78}{\scriptsize$\pm$.03} & 0.52{\scriptsize$\pm$.01} & \textbf{0.87}{\scriptsize$\pm$.02} & 0.62{\scriptsize$\pm$.01} & \textbf{0.87}{\scriptsize$\pm$.02} & 0.62{\scriptsize$\pm$.02} & \textbf{0.93}{\scriptsize$\pm$.02} & 0.80{\scriptsize$\pm$.02} \\
Scene & \textbf{0.59}{\scriptsize$\pm$.01} & 0.51{\scriptsize$\pm$.00} & \textbf{0.51}{\scriptsize$\pm$.01} & 0.50{\scriptsize$\pm$.00} & \textbf{0.54}{\scriptsize$\pm$.02} & 0.50{\scriptsize$\pm$.00} & \textbf{0.50}{\scriptsize$\pm$.00} & 0.50{\scriptsize$\pm$.00} & 0.49{\scriptsize$\pm$.01} & \textbf{0.50}{\scriptsize$\pm$.00} \\
Cube double & \textbf{0.62}{\scriptsize$\pm$.02} & 0.52{\scriptsize$\pm$.01} & \textbf{0.80}{\scriptsize$\pm$.03} & 0.50{\scriptsize$\pm$.00} & \textbf{0.78}{\scriptsize$\pm$.03} & 0.50{\scriptsize$\pm$.01} & 0.48{\scriptsize$\pm$.01} & \textbf{0.49}{\scriptsize$\pm$.00} & \textbf{0.69}{\scriptsize$\pm$.03} & 0.51{\scriptsize$\pm$.01} \\
\bottomrule
\end{tabular}
\end{table*}

\begin{table}[!h]
\centering
\footnotesize\setlength{\tabcolsep}{3.5pt}
\caption{\textbf{Data of Fig.~\ref{fig:viol_heatmap}, shifts.} AUROC of $s^2$ for shifted against clean windows, $\pm$ bootstrap sd over windows (paired); bold = better model. vis. = visual, phy. = physics shift.}
\label{tab:shift}
\begin{tabular}{llcc}
\toprule
Environment & shift & LeWM & Dreamer \\
\midrule
TwoRoom & colour (vis.) & \textbf{1.00}{\scriptsize$\pm$.00} & 0.75{\scriptsize$\pm$.05} \\
 & background (vis.) & \textbf{1.00}{\scriptsize$\pm$.00} & 0.51{\scriptsize$\pm$.00} \\
 & speed (phy.) & 0.69{\scriptsize$\pm$.04} & \textbf{0.79}{\scriptsize$\pm$.04} \\
\addlinespace[2pt]
PointMaze & colour (vis.) & \textbf{1.00}{\scriptsize$\pm$.00} & 0.74{\scriptsize$\pm$.03} \\
 & floor (vis.) & \textbf{1.00}{\scriptsize$\pm$.00} & 0.54{\scriptsize$\pm$.01} \\
 & light (vis.) & \textbf{1.00}{\scriptsize$\pm$.00} & 0.52{\scriptsize$\pm$.01} \\
\addlinespace[2pt]
Reacher & colour (vis.) & \textbf{1.00}{\scriptsize$\pm$.00} & 0.86{\scriptsize$\pm$.03} \\
 & light (vis.) & \textbf{1.00}{\scriptsize$\pm$.00} & 0.52{\scriptsize$\pm$.01} \\
 & density (phy.) & 0.50{\scriptsize$\pm$.01} & \textbf{0.54}{\scriptsize$\pm$.01} \\
\addlinespace[2pt]
PushT & colour (vis.) & \textbf{0.93}{\scriptsize$\pm$.02} & 0.76{\scriptsize$\pm$.04} \\
 & background (vis.) & \textbf{0.96}{\scriptsize$\pm$.02} & 0.79{\scriptsize$\pm$.04} \\
 & size (phy.) & \textbf{0.60}{\scriptsize$\pm$.03} & 0.52{\scriptsize$\pm$.01} \\
\addlinespace[2pt]
Cube & colour (vis.) & \textbf{0.83}{\scriptsize$\pm$.03} & 0.51{\scriptsize$\pm$.01} \\
 & light (vis.) & \textbf{0.99}{\scriptsize$\pm$.00} & 0.54{\scriptsize$\pm$.02} \\
 & camera (vis.) & \textbf{0.83}{\scriptsize$\pm$.03} & 0.68{\scriptsize$\pm$.03} \\
 & size (phy.) & \textbf{0.56}{\scriptsize$\pm$.02} & 0.50{\scriptsize$\pm$.01} \\
\addlinespace[2pt]
Ball-in-cup & colour (vis.) & \textbf{0.90}{\scriptsize$\pm$.03} & 0.56{\scriptsize$\pm$.01} \\
 & light (vis.) & \textbf{1.00}{\scriptsize$\pm$.00} & 0.50{\scriptsize$\pm$.00} \\
 & gravity (phy.) & \textbf{0.78}{\scriptsize$\pm$.03} & 0.69{\scriptsize$\pm$.02} \\
 & density (phy.) & 0.60{\scriptsize$\pm$.03} & \textbf{0.61}{\scriptsize$\pm$.02} \\
\addlinespace[2pt]
Finger & colour (vis.) & \textbf{0.89}{\scriptsize$\pm$.02} & 0.50{\scriptsize$\pm$.01} \\
 & light (vis.) & \textbf{0.67}{\scriptsize$\pm$.02} & 0.49{\scriptsize$\pm$.01} \\
 & friction (phy.) & \textbf{0.99}{\scriptsize$\pm$.01} & 0.87{\scriptsize$\pm$.03} \\
 & density (phy.) & \textbf{0.95}{\scriptsize$\pm$.02} & 0.82{\scriptsize$\pm$.03} \\
\addlinespace[2pt]
Fetch push & colour (vis.) & \textbf{0.90}{\scriptsize$\pm$.03} & 0.50{\scriptsize$\pm$.01} \\
 & light (vis.) & \textbf{1.00}{\scriptsize$\pm$.00} & 0.52{\scriptsize$\pm$.01} \\
\addlinespace[2pt]
Scene & colour (vis.) & 0.51{\scriptsize$\pm$.01} & \textbf{0.52}{\scriptsize$\pm$.01} \\
 & light (vis.) & 0.49{\scriptsize$\pm$.02} & \textbf{0.51}{\scriptsize$\pm$.01} \\
 & camera (vis.) & \textbf{0.92}{\scriptsize$\pm$.03} & 0.58{\scriptsize$\pm$.02} \\
 & size (phy.) & 0.48{\scriptsize$\pm$.01} & \textbf{0.50}{\scriptsize$\pm$.01} \\
\addlinespace[2pt]
Cube double & colour (vis.) & \textbf{0.70}{\scriptsize$\pm$.04} & 0.49{\scriptsize$\pm$.01} \\
 & light (vis.) & 0.42{\scriptsize$\pm$.02} & \textbf{0.50}{\scriptsize$\pm$.01} \\
 & camera (vis.) & 0.51{\scriptsize$\pm$.03} & \textbf{0.82}{\scriptsize$\pm$.03} \\
 & size (phy.) & \textbf{0.64}{\scriptsize$\pm$.04} & 0.50{\scriptsize$\pm$.01} \\
\bottomrule
\end{tabular}
\end{table}

\subsection*{Uncertainty extraction at Dreamers posterior features}
\label{app:posterior}

The record used throughout the paper reads the input of the predictor's last linear layer, $\boldsymbol{\psi}_{t+1} = \mathrm{prior\_features}(\mathbf h_{t+1})$, the feature of the RSSM's prior head. This is the natural place for a record: it is the last stage before the prediction and the only feature that exists along an imagined rollout. The stage trace of Sec.~\ref{sec:analyitcal_framework}, however, locates Dreamer's attenuation of post-hoc unceratainty one stage earlier, at the GRU update that maps the posterior of the current frame to $\mathbf h_{t+1}$. As a control we therefore fit the same Gram head to the posterior feature $\boldsymbol{\psi}^{\mathrm{post}}_t$, the input of the posterior head's last linear layer, $[\mathbf h_t, \mathbf z_t]$ passed through its hidden layer, on the same training windows, and score the posterior feature of the last context frame. 
Reading the posterior removes the GRU attenuation from the chain, and the signal recovers to the extent that this attenuation was the cause. Single-frame violations are detected at LeWM's level in every environment but Cube (Table~\ref{tab:post_violations}). The dynamics and viewpoint shifts, speed, gravity, density, friction, size and camera, rise from near chance to $0.6$--$1.0$ (Table~\ref{tab:post_shift}). Lighting shifts recover only partially, and unfamiliar-state detection and error ranking, which ask the record about the model's training support and its own predictions rather than about the current frame, change by at most a few points (Tables~\ref{tab:post_detection} and \ref{tab:post_error}). The posterior feature requires the observed frame and is not available in imagination, so this record cannot replace the prior head for planning.

\begin{table}[!h]
\centering
\small
\caption{\textbf{Unfamiliar-state detection with Dreamer's posterior head.} AUROC for recognising held-out states of a withheld region (mean $\pm$ sd over quadrants). Dreamer columns: the paper's record on the prior feature $\boldsymbol{\psi}_{t+1}$ and the Laplace head fitted to the posterior feature of the current frame, both refitted on the same rows; LeWM as in the main text. Bold = best.}
\label{tab:post_detection}
\begin{tabular}{llccc}
\toprule
& & LeWM & \multicolumn{2}{c}{Dreamer} \\ \cmidrule(lr){4-5}
Environment & factor & $s^2$ & prior & posterior \\
\midrule
TwoRoom & body & 0.96{\scriptsize$\pm$.02} & 0.95{\scriptsize$\pm$.02} & \textbf{0.99}{\scriptsize$\pm$.00} \\
PointMaze & body & \textbf{1.00}{\scriptsize$\pm$.00} & 0.98{\scriptsize$\pm$.02} & 0.99{\scriptsize$\pm$.01} \\
Reacher & body & \textbf{0.84}{\scriptsize$\pm$.02} & 0.80{\scriptsize$\pm$.02} & 0.83{\scriptsize$\pm$.02} \\
PushT & body & \textbf{0.96}{\scriptsize$\pm$.01} & 0.80{\scriptsize$\pm$.04} & 0.84{\scriptsize$\pm$.03} \\
PushT & object & \textbf{0.98}{\scriptsize$\pm$.01} & 0.81{\scriptsize$\pm$.05} & 0.81{\scriptsize$\pm$.06} \\
Cube & body & 0.97{\scriptsize$\pm$.00} & 0.97{\scriptsize$\pm$.01} & \textbf{0.98}{\scriptsize$\pm$.01} \\
Cube & object & \textbf{0.99}{\scriptsize$\pm$.00} & 0.94{\scriptsize$\pm$.01} & 0.95{\scriptsize$\pm$.01} \\
Ball-in-cup & body & \textbf{0.98}{\scriptsize$\pm$.01} & 0.95{\scriptsize$\pm$.02} & 0.97{\scriptsize$\pm$.01} \\
Ball-in-cup & object & \textbf{0.98}{\scriptsize$\pm$.00} & 0.93{\scriptsize$\pm$.03} & 0.96{\scriptsize$\pm$.02} \\
Finger & body & 0.81{\scriptsize$\pm$.10} & 0.88{\scriptsize$\pm$.07} & \textbf{0.91}{\scriptsize$\pm$.06} \\
Finger & object & 0.78{\scriptsize$\pm$.03} & 0.84{\scriptsize$\pm$.00} & \textbf{0.84}{\scriptsize$\pm$.02} \\
Fetch push & body & \textbf{0.97}{\scriptsize$\pm$.00} & 0.89{\scriptsize$\pm$.03} & 0.90{\scriptsize$\pm$.03} \\
Fetch push & object & \textbf{0.97}{\scriptsize$\pm$.00} & 0.90{\scriptsize$\pm$.03} & 0.92{\scriptsize$\pm$.02} \\
\bottomrule
\end{tabular}
\end{table}

\begin{table*}[h]
\centering
\small
\caption{\textbf{Error ranking with Dreamer's posterior head.} Spearman correlation between the record and Dreamer's own one-step error (posterior--prior KL), $\pm$ block-bootstrap sd. Prior: record scored against the KL of its own step; posterior (next): the posterior feature of frame $t$ against the KL of step $t+1$; posterior (same): against the KL of step $t$. LeWM as in the main text. Bold = best.}
\label{tab:post_error}
\begin{tabular}{lcccc}
\toprule
& LeWM & \multicolumn{3}{c}{Dreamer} \\ \cmidrule(lr){3-5}
Environment & $s^2$ & prior & posterior (next) & posterior (same) \\
\midrule
TwoRoom & \textbf{0.36}{\scriptsize$\pm$.01} & 0.32{\scriptsize$\pm$.01} & 0.20{\scriptsize$\pm$.01} & 0.28{\scriptsize$\pm$.01} \\
PointMaze & \textbf{0.62}{\scriptsize$\pm$.01} & 0.16{\scriptsize$\pm$.01} & 0.19{\scriptsize$\pm$.01} & 0.21{\scriptsize$\pm$.01} \\
Reacher & \textbf{0.45}{\scriptsize$\pm$.01} & 0.19{\scriptsize$\pm$.01} & 0.13{\scriptsize$\pm$.01} & 0.19{\scriptsize$\pm$.01} \\
PushT & 0.45{\scriptsize$\pm$.01} & 0.52{\scriptsize$\pm$.01} & 0.49{\scriptsize$\pm$.01} & \textbf{0.53}{\scriptsize$\pm$.01} \\
Cube & \textbf{0.28}{\scriptsize$\pm$.01} & 0.22{\scriptsize$\pm$.01} & 0.21{\scriptsize$\pm$.01} & 0.24{\scriptsize$\pm$.01} \\
Ball-in-cup & \textbf{0.59}{\scriptsize$\pm$.01} & 0.39{\scriptsize$\pm$.01} & 0.40{\scriptsize$\pm$.01} & 0.41{\scriptsize$\pm$.01} \\
Finger & \textbf{0.37}{\scriptsize$\pm$.01} & 0.23{\scriptsize$\pm$.01} & 0.17{\scriptsize$\pm$.01} & 0.32{\scriptsize$\pm$.01} \\
Fetch push & 0.05{\scriptsize$\pm$.01} & \textbf{0.28}{\scriptsize$\pm$.01} & 0.14{\scriptsize$\pm$.01} & 0.21{\scriptsize$\pm$.01} \\
\bottomrule
\end{tabular}
\end{table*}

\begin{figure}[!h]
\centering
\includegraphics[width=\linewidth]{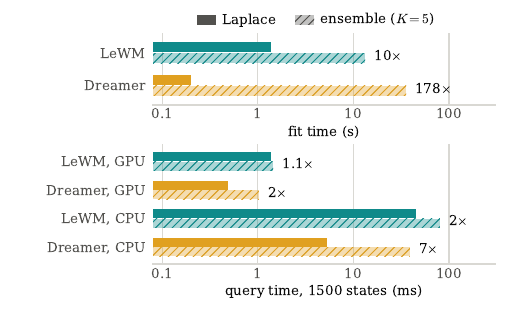}
\caption{\textbf{Computational cost.} Wall-clock time of fitting the Laplace record (filled) and the $K=5$ ensemble (hatched), and of one query of 1500 states on GPU and CPU; one H200, numbers give the ensemble-to-Laplace ratio.}
\label{fig:cost}
\end{figure}

\subsection*{Additional Results Planning Tasks}
\label{app:planning_base}

Tables~\ref{tab:planning_std} and \ref{tab:planning_base} give the success rates of the frozen world models under CEM planning without uncertainty penalty, which are the baselines against which the deltas of figure.~\ref{fig:planning_combo} are measured. Table~\ref{tab:planning_std} follows the standard evaluation protocol of LeWM, random starts with a goal 25 steps ahead and a budget of 50 steps, and shows that both models solve the tasks at comparable rates in most environments; Dreamer trails on PushT, LeWM on TwoRoom and Ball-in-cup. Table~\ref{tab:planning_base} gives the same quantity for the two settings in which the penalty is applied, starts in the rare region of the training data (P1) the training data (P1) and a three-fold longer goal distance and a three-fold longer goal distance and budget (P2). Both settings lower the success rates considerably.

\begin{table*}[h]
\centering
\footnotesize\setlength{\tabcolsep}{4pt}
\newcommand{\e}[2]{#1{\scriptsize$\pm$.#2}}
\newcommand{\be}[2]{\textbf{#1}{\scriptsize$\pm$.#2}}
\caption{\textbf{Physical-violation detection with Dreamer's posterior head.} AUROC of the record for a violated against the clean window, $\pm$ bootstrap sd over base states (paired): L = LeWM $s^2$, D = Dreamer on the prior feature, D$_{\mathrm{post}}$ = Dreamer's Laplace head on the posterior feature of the last context frame (prior-head $\sigma^2,\tau^2$). Teleport, recolour, removal and interpenetration in the current frame, reversal over the last eight frames. Bold = best; \mbox{--} = not applicable.}
\label{tab:post_violations}
\begin{tabular}{llcccccccc}
\toprule
Violation & & TwoRoom & PointMaze & Reacher & PushT & Cube & Ball-in-cup & Finger & Fetch push \\
\midrule
teleport  & L & \e{0.71}{03} & \e{0.99}{00} & \e{0.97}{01} & \e{0.87}{03} & \be{0.90}{02} & \e{1.00}{00} & \e{0.98}{01} & \e{0.88}{02} \\
          & D & \e{0.69}{02} & \e{0.73}{02} & \e{0.77}{02} & \e{0.58}{01} & \e{0.51}{01} & \e{0.61}{02} & \e{0.74}{03} & \e{0.66}{02} \\
          & D$_{\mathrm{post}}$ & \be{1.00}{00} & \be{1.00}{00} & \be{1.00}{00} & \be{0.99}{01} & \e{0.84}{03} & \be{1.00}{00} & \be{1.00}{00} & \be{0.99}{00} \\
\midrule
recolour  & L & \be{1.00}{00} & \e{1.00}{00} & \be{1.00}{00} & \e{0.94}{02} & \e{0.80}{03} & \e{0.91}{02} & \e{0.84}{02} & \be{0.78}{03} \\
          & D & \e{0.65}{02} & \e{0.70}{02} & \e{0.77}{02} & \e{0.63}{02} & \e{0.50}{00} & \e{0.56}{01} & \e{0.50}{01} & \e{0.52}{01} \\
          & D$_{\mathrm{post}}$ & \be{1.00}{00} & \be{1.00}{00} & \be{1.00}{00} & \be{0.94}{02} & \be{0.81}{02} & \be{0.99}{00} & \be{0.88}{02} & \e{0.78}{02} \\
\midrule
removal   & L & \be{1.00}{00} & \be{1.00}{00} & \be{1.00}{00} & \e{0.97}{01} & \be{0.75}{03} & \e{0.98}{01} & \e{0.92}{02} & \e{0.87}{02} \\
          & D & \e{0.64}{02} & \e{0.71}{02} & \e{0.78}{02} & \e{0.62}{01} & \e{0.50}{00} & \e{0.59}{01} & \e{0.55}{01} & \e{0.62}{01} \\
          & D$_{\mathrm{post}}$ & \be{1.00}{00} & \be{1.00}{00} & \be{1.00}{00} & \be{0.97}{01} & \e{0.67}{02} & \be{1.00}{00} & \be{0.96}{02} & \be{0.95}{02} \\
\midrule
interpen. & L & -- & -- & \be{1.00}{00} & \e{0.77}{03} & \be{0.50}{00} & \e{1.00}{00} & \e{0.50}{00} & \e{0.87}{02} \\
          & D & -- & -- & \e{0.71}{02} & \e{0.62}{02} & \e{0.50}{00} & \e{0.58}{01} & \e{0.50}{00} & \e{0.62}{02} \\
          & D$_{\mathrm{post}}$ & -- & -- & \e{1.00}{00} & \be{0.95}{01} & \e{0.50}{01} & \be{1.00}{00} & \be{0.50}{00} & \be{0.97}{01} \\
\midrule
reversed  & L & -- & -- & \e{0.89}{03} & \e{0.80}{03} & \be{0.85}{02} & \be{1.00}{00} & \e{0.89}{02} & \e{0.93}{02} \\
          & D & -- & -- & \e{0.95}{02} & \e{0.62}{03} & \e{0.50}{00} & \e{0.81}{02} & \e{0.72}{02} & \e{0.80}{02} \\
          & D$_{\mathrm{post}}$ & -- & -- & \be{0.99}{01} & \be{0.80}{03} & \e{0.79}{03} & \e{0.99}{01} & \be{0.92}{02} & \be{0.99}{00} \\
\bottomrule
\end{tabular}
\end{table*}

\begin{table*}[h]
\centering
\footnotesize\setlength{\tabcolsep}{3.5pt}
\caption{\textbf{Consistent shifts with Dreamer's posterior head.} AUROC of the record for shifted against clean windows, $\pm$ bootstrap sd over windows (paired); columns as in Table~\ref{tab:post_violations}. vis. = visual, phy. = physics shift. Bold = best.}
\label{tab:post_shift}
\begin{tabular}{llccc}
\toprule
& & LeWM & \multicolumn{2}{c}{Dreamer} \\ \cmidrule(lr){4-5}
Environment & shift & $s^2$ & prior & posterior \\
\midrule
TwoRoom & colour (vis.) & \textbf{1.00}{\scriptsize$\pm$.00} & 0.75{\scriptsize$\pm$.05} & \textbf{1.00}{\scriptsize$\pm$.00} \\
 & background (vis.) & \textbf{1.00}{\scriptsize$\pm$.00} & 0.51{\scriptsize$\pm$.00} & 0.61{\scriptsize$\pm$.01} \\
 & speed (phy.) & 0.69{\scriptsize$\pm$.04} & 0.79{\scriptsize$\pm$.04} & \textbf{1.00}{\scriptsize$\pm$.00} \\
\addlinespace[2pt]
PointMaze & colour (vis.) & \textbf{1.00}{\scriptsize$\pm$.00} & 0.74{\scriptsize$\pm$.03} & 0.99{\scriptsize$\pm$.00} \\
 & floor (vis.) & \textbf{1.00}{\scriptsize$\pm$.00} & 0.54{\scriptsize$\pm$.01} & 0.96{\scriptsize$\pm$.02} \\
 & light (vis.) & \textbf{1.00}{\scriptsize$\pm$.00} & 0.52{\scriptsize$\pm$.01} & 0.62{\scriptsize$\pm$.02} \\
\addlinespace[2pt]
Reacher & colour (vis.) & \textbf{1.00}{\scriptsize$\pm$.00} & 0.86{\scriptsize$\pm$.03} & \textbf{1.00}{\scriptsize$\pm$.00} \\
 & light (vis.) & \textbf{1.00}{\scriptsize$\pm$.00} & 0.52{\scriptsize$\pm$.01} & 0.67{\scriptsize$\pm$.02} \\
 & density (phy.) & 0.50{\scriptsize$\pm$.01} & 0.54{\scriptsize$\pm$.01} & \textbf{0.58}{\scriptsize$\pm$.01} \\
\addlinespace[2pt]
PushT & colour (vis.) & \textbf{0.93}{\scriptsize$\pm$.02} & 0.76{\scriptsize$\pm$.04} & 0.69{\scriptsize$\pm$.05} \\
 & background (vis.) & 0.96{\scriptsize$\pm$.02} & 0.79{\scriptsize$\pm$.04} & \textbf{1.00}{\scriptsize$\pm$.00} \\
 & size (phy.) & 0.60{\scriptsize$\pm$.03} & 0.52{\scriptsize$\pm$.01} & \textbf{0.63}{\scriptsize$\pm$.02} \\
\addlinespace[2pt]
Cube & colour (vis.) & 0.83{\scriptsize$\pm$.03} & 0.51{\scriptsize$\pm$.01} & \textbf{0.83}{\scriptsize$\pm$.03} \\
 & light (vis.) & 0.99{\scriptsize$\pm$.00} & 0.54{\scriptsize$\pm$.02} & \textbf{1.00}{\scriptsize$\pm$.00} \\
 & camera (vis.) & 0.83{\scriptsize$\pm$.03} & 0.68{\scriptsize$\pm$.03} & \textbf{1.00}{\scriptsize$\pm$.00} \\
 & size (phy.) & 0.56{\scriptsize$\pm$.02} & 0.50{\scriptsize$\pm$.01} & \textbf{0.69}{\scriptsize$\pm$.02} \\
\addlinespace[2pt]
Ball-in-cup & colour (vis.) & 0.90{\scriptsize$\pm$.03} & 0.56{\scriptsize$\pm$.01} & \textbf{0.93}{\scriptsize$\pm$.02} \\
 & light (vis.) & \textbf{1.00}{\scriptsize$\pm$.00} & 0.50{\scriptsize$\pm$.00} & 0.53{\scriptsize$\pm$.01} \\
 & gravity (phy.) & 0.78{\scriptsize$\pm$.03} & 0.69{\scriptsize$\pm$.02} & \textbf{0.98}{\scriptsize$\pm$.01} \\
 & density (phy.) & 0.60{\scriptsize$\pm$.03} & 0.61{\scriptsize$\pm$.02} & \textbf{0.81}{\scriptsize$\pm$.02} \\
\addlinespace[2pt]
Finger & colour (vis.) & \textbf{0.89}{\scriptsize$\pm$.02} & 0.50{\scriptsize$\pm$.01} & 0.85{\scriptsize$\pm$.03} \\
 & light (vis.) & \textbf{0.67}{\scriptsize$\pm$.02} & 0.49{\scriptsize$\pm$.01} & 0.58{\scriptsize$\pm$.02} \\
 & friction (phy.) & 0.99{\scriptsize$\pm$.01} & 0.87{\scriptsize$\pm$.03} & \textbf{0.99}{\scriptsize$\pm$.00} \\
 & density (phy.) & 0.95{\scriptsize$\pm$.02} & 0.82{\scriptsize$\pm$.03} & \textbf{0.97}{\scriptsize$\pm$.01} \\
\addlinespace[2pt]
Fetch push & colour (vis.) & \textbf{0.90}{\scriptsize$\pm$.03} & 0.50{\scriptsize$\pm$.01} & 0.68{\scriptsize$\pm$.02} \\
 & light (vis.) & \textbf{1.00}{\scriptsize$\pm$.00} & 0.52{\scriptsize$\pm$.01} & 0.73{\scriptsize$\pm$.02} \\
\bottomrule
\end{tabular}
\end{table*}

\begin{table}[h]
\centering
\small
\caption{\textbf{Planning success rates.} Success rate (\%) of CEM planning with the frozen world model and no penalty under the evaluation protocol of LeWM (random starts, goal offset 25 steps, budget 50 steps); $\pm$ binomial se over episodes. }
\label{tab:planning_std}
\begin{tabular}{lcc}
\toprule
Environment & LeWM & Dreamer \\
\midrule
TwoRoom & 85.0{\scriptsize$\pm$2.5} & 91.5{\scriptsize$\pm$2.0} \\
PointMaze & 91.0{\scriptsize$\pm$2.0} & 99.5{\scriptsize$\pm$0.5} \\
Reacher & 37.5{\scriptsize$\pm$3.4} & 40.5{\scriptsize$\pm$3.4} \\
PushT & 83.0{\scriptsize$\pm$2.6} & 30.5{\scriptsize$\pm$3.3} \\
Cube & 64.0{\scriptsize$\pm$3.3} & 60.5{\scriptsize$\pm$3.4} \\
Ball-in-cup & 60.5{\scriptsize$\pm$3.5} & 71.0{\scriptsize$\pm$3.2} \\
Finger$^{\dagger}$ & 7.5{\scriptsize$\pm$1.8} & 7.0{\scriptsize$\pm$1.8} \\
Fetch push & 48.0{\scriptsize$\pm$3.5} & 46.5{\scriptsize$\pm$3.6} \\
Scene & 65.5{\scriptsize$\pm$3.4} & 60.5{\scriptsize$\pm$3.5} \\
Cube double & 84.5{\scriptsize$\pm$2.6} & 58.5{\scriptsize$\pm$3.5} \\
\bottomrule
\end{tabular}
\end{table}

\begin{table}[!h]
\centering
\small
\caption{\textbf{Planning success rates rare region starts and long horizon goals.} Success rate (\%) of CEM planning with the frozen world model under rare-region starts (P1) and the long horizon (P2); $\pm$ bootstrap se over episodes.}
\label{tab:planning_base}
\begin{tabular}{lcccc}
\toprule
& \multicolumn{2}{c}{P1: rare-region starts} & \multicolumn{2}{c}{P2: long horizon} \\ \cmidrule(lr){2-3} \cmidrule(lr){4-5}
Environment & LeWM & Dreamer & LeWM & Dreamer \\
\midrule
TwoRoom & 76.8{\scriptsize$\pm$1.8} & 89.5{\scriptsize$\pm$1.5} & 34.2{\scriptsize$\pm$2.4} & 30.0{\scriptsize$\pm$2.3} \\
PointMaze & 96.3{\scriptsize$\pm$0.8} & 99.0{\scriptsize$\pm$0.4} & 21.3{\scriptsize$\pm$1.7} & 60.3{\scriptsize$\pm$2.0} \\
Reacher & 15.0{\scriptsize$\pm$1.4} & 15.5{\scriptsize$\pm$1.5} & 54.3{\scriptsize$\pm$2.0} & 59.7{\scriptsize$\pm$2.0} \\
PushT & 73.5{\scriptsize$\pm$2.2} & 24.2{\scriptsize$\pm$2.1} & 19.5{\scriptsize$\pm$2.0} & 2.5{\scriptsize$\pm$0.8} \\
Cube & 70.0{\scriptsize$\pm$2.2} & 56.0{\scriptsize$\pm$2.4} & 65.8{\scriptsize$\pm$2.4} & 64.8{\scriptsize$\pm$2.4} \\
Ball-in-cup & 44.3{\scriptsize$\pm$2.1} & 42.2{\scriptsize$\pm$2.0} & 17.0{\scriptsize$\pm$2.7} & 28.0{\scriptsize$\pm$3.1} \\
Finger$^{\dagger}$ & 0.5{\scriptsize$\pm$0.5} & 0.0{\scriptsize$\pm$0.0} & 5.5{\scriptsize$\pm$1.6} & 7.5{\scriptsize$\pm$1.9} \\
Fetch push & 42.8{\scriptsize$\pm$2.0} & 44.2{\scriptsize$\pm$2.1} & 9.2{\scriptsize$\pm$1.2} & 9.0{\scriptsize$\pm$1.2} \\
Scene & 25.2{\scriptsize$\pm$1.8} & 29.7{\scriptsize$\pm$1.9} & 47.2{\scriptsize$\pm$2.0} & 38.3{\scriptsize$\pm$2.0} \\
Cube double & 85.2{\scriptsize$\pm$1.4} & 56.0{\scriptsize$\pm$2.0} & 19.2{\scriptsize$\pm$1.6} & 2.3{\scriptsize$\pm$0.6} \\
\bottomrule
\end{tabular}
\end{table}

\subsection*{Estimator Ablation with Deep Ensemble}
\label{app:ensemble}

\begin{figure}[h]
\centering
\includegraphics[width=\linewidth]{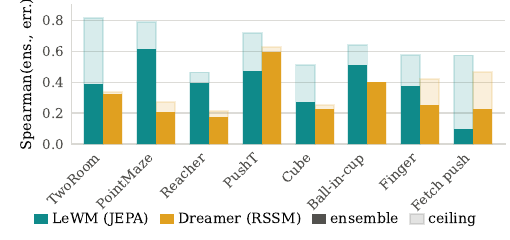}
\caption{\textbf{Error ranking, post-hoc ensemble.} As Fig.~\ref{fig:error_ranking}, with the ensemble disagreement in place of $s^2$.}
\label{fig:error_ranking_ens}
\end{figure}

To check that the comparison does not depend on the Laplace estimator, we repeat downstream tasks with a post-hoc ensemble on the same frozen models and the same features: $M=5$ MLP heads trained on bootstrap resamples of the training transitions, with the cross-head prediction variance (LeWM) or the Jensen--Shannon disagreement of the categorical prior (Dreamer) in place of $s^2$. Figures~\ref{fig:viol_heatmap_ens} and \ref{fig:error_ranking_ens} reproduce Figs.~\ref{fig:viol_heatmap} and \ref{fig:error_ranking} with the ensemble on the eight environments of the main comparison. The two estimators agree almost perfectly: the cells at chance stay at chance, the ranking of the two models is unchanged. The paradigm difference is a property of the features the world models expose, not of the estimator.

\subsection*{Computational comparison of Last Layer Laplace and Ensemble}
\label{app:cost}

Both estimators operate on the last-layer features $\boldsymbol{\psi}\in\mathbb{R}^{F}$ of the frozen predictor ($F=2048$ for LeWM, $512$ for Dreamer). Let $N$ be the number of training features, $H=512$ the width of an ensemble head, $D$ its output dimension ($192$ for LeWM, $32\times32$ for Dreamer), $K=5$ the number of heads and $E=20$ the number of epochs. Fitting the Laplace record costs $\mathcal{O}(NF^2+F^3)$, one pass for the Gram matrix and one eigendecomposition; a query is a quadratic form, $\mathcal{O}(F^2)$ per state. Training the ensemble costs $\mathcal{O}(KEN(FH+HD))$ and a query $\mathcal{O}(K(FH+HD))$. For the dimensions used here the Laplace fit is $80\times$ (LeWM) and $900\times$ (Dreamer) cheaper; a query costs the same for LeWM and is $15\times$ cheaper for Dreamer. Figure~\ref{fig:cost} reports the measured wall-clock times on the paper's feature sets (one H200 GPU; queries also on 8 CPU threads). Both estimators are negligible against the forward pass of the world model in our setting; however in large scale settings this could become of significant impact. The advantage of the closed-form record is the fit, which has no training loop and no hyper-parameters beyond $(\sigma^2,\tau^2)$.

\section*{D Other}
\label{app:landscapes}

\subsection*{Uncertainty Map Computation}
\label{app:maps}

This section details how the uncertainty maps of Fig.~\ref{fig:landscapes} were computed. Let $s_t$ be the recorded simulator state of a held-out frame and $q \subset s_t$ the position of the controlled entity (agent, fingertip or object). For each point $g$ of a grid over the entity's workspace we form the counterfactual state $s_t^{(g)}$ by setting $q \leftarrow g$ with the entity's velocities set to zero; for the arms, $g$ is converted to joint angles by inverse kinematics, taking the configuration most frequent in the training data. The map value is the Laplace record of the resulting feature,
\begin{equation}
  u(g) \;=\; \phi\!\left(s_t^{(g)}\right)^{\!\top} \Sigma\, \phi\!\left(s_t^{(g)}\right),
\end{equation}
where $\phi$ is the model's last-layer feature (LeWM: the predictor's input feature at the current step; Dreamer: the posterior feature, averaged over eight samples of the stochastic state) and $\Sigma$ is the Laplace covariance fitted on the training set. Panels show $\log_2\big(u(g)/\tilde u\big)$, with $\tilde u$ the median over physically possible grid points. Physically impossible positions (inside walls, beyond the table edge, outside the string's reach or the arm's joint limits) are rendered without simulating contact, so any increase in uncertainty there reflects the model's representation of where the entity can be.

\subsection*{Gradient Visualizations of Uncertainty}

\label{app:maps}
We provide additional examples of gradient maps that show which pixels of the current frame affect the uncertainty of the next-step prediction. We attribute $s^2$ to the pixels of the current frame $\mathbf{x}_H$ with SmoothGrad,
\begin{align}
S(p) &= \frac{1}{M}\sum_{m=1}^{M}\,\sum_{c}\Bigl|\frac{\partial\, s^2(\mathbf{x}_{1:H-1},\,\mathbf{x}_H+\boldsymbol{\epsilon}_m)}{\partial\, x_{H,c,p}}\Bigr|, \notag\\
\boldsymbol{\epsilon}_m &\sim \mathcal{N}\bigl(0,\,(0.15\,\sigma_{\mathrm{in}})^2\mathbf{I}\bigr),
\label{eq:smoothgrad}
\end{align}
with $M=8$ draws ($\boldsymbol{\epsilon}_1=0$), $c$ the colour channels, $p$ the pixel and $\sigma_{\mathrm{in}}$ the per-channel standard deviation of the normalised input. The gradient is taken through the full chain from pixels to $s^2$; for Dreamer, $s^2$ is averaged over four posterior samples and the $64\times64$ map is upsampled to the frame. Each map is shown on a dimmed copy of the frame and scaled to its own 99th percentile, so it shows \emph{where} the uncertainty is sensitive, not how much it changes. Figures~\ref{fig:maps_a} and \ref{fig:maps_b} show one held-out window per environment, for the clean frame and for the teleport, recolour and removal violations of the object (of the body where the environment has no object). A quantitative analysis of these maps is beyond the scope of this paper, and the examples shown are single windows that need not be representative. With this caveat, the examples are consistent with our diagnosis. LeWM's sensitivity sits on the manipulated entity (the cube, the puck, the T block, the ball) and on the region the entity can reach, such as the disc swept by the Reacher arm, the maze corridors or the table in front of the gripper, and it moves with the entity when the entity is teleported or recoloured; the maps suggest, though we do not claim, that the record responds to the physically relevant part of the frame. Dreamer's sensitivity is dominated by the arm and the large surfaces it reconstructs and changes little with the violation; on the body-only environments it concentrates on the agent, on the object environments it is diffuse and partly erratic, and the manipulated object receives little or no mass even when it is the entity that was violated.

\end{document}